\ifx\varDocClass\undefined
\def \varDocClass{IEEEtran}
\fi

\ifx\varKeyvals\undefined
\def \varKeyvals{conference}
\fi

\def\varAuthors{}
\def\varInstitutions{}

\makeatletter
\newcommand{\addAuthor}[2]{
	\g@addto@macro\varAuthors{#1}
	\@for\varRef:={#2}\do{
		\expandafter\g@addto@macro\expandafter\varAuthors\expandafter{\expandafter\IEEEauthorrefmark\expandafter{\varRef}}
	}
	\g@addto@macro\varAuthors{ }
}
\newcommand{\addInst}[2]{%
	\g@addto@macro\varInstitutions{ %
		\IEEEauthorblockA{\IEEEauthorrefmark{#1}#2}
	}
}

\newcounter{affId}
\newcommand{\addInstitution}[4]{%
	\stepcounter{affId}
	\edef\myval{\theaffId}
	\expandafter\addInst\expandafter{\myval}{#1, #2, #3\\#4}
}
\makeatother

\newcommand{\addAbstract}[1]{\def\varAbstract{#1}}
\newcommand{\addTitle}[1]{\def\varTitle{#1}}
\newcommand{\addKeywords}[1]{\def\varKeywords{#1}}
\newcommand{\setRunningHeads}[1]{}

\newcommand{\insertTopmatter}{
	\title{\varTitle}
	\author{\IEEEauthorblockN{\varAuthors}\varInstitutions}
	\maketitle
	\begin{abstract}
		\varAbstract
	\end{abstract}
	\begin{IEEEkeywords}
		\varKeywords
	\end{IEEEkeywords}
}

\newcommand{\setBibStyle}{\bibliographystyle{bibliography/IEEEtran}}

\addTitle{Guided Image Feature Matching using Feature Spatial Order}

\addAuthor{Chin-Hung Teng}{1}
\addAuthor{Ben-Jian Dong}{2}
\addInstitution{Department of Information Communication, Yuan Ze University}{Taoyuan}{Taiwan}{chteng@saturn.yzu.edu.tw}
\addInstitution{School of Artificial Intelligence and Electrical Engineering, Guizhou Institute of Technology}{Guizhou}{China}{idbj.real@foxmail.com}

\addAbstract{
	Image feature matching plays a vital role in many computer vision tasks. Although many image feature detection and matching techniques have been proposed over the past few decades, it is still time-consuming to match feature points in two images, especially for images with a large number of detected features. Feature spatial order can estimate the probability that a pair of features is correct. Since it is a completely independent concept from epipolar geometry, it can be used to complement epipolar geometry in guiding feature match in a target region so as to improve matching efficiency. In this paper, we integrate the concept of feature spatial order into a progressive matching framework. We use some of the initially matched features to build a computational model of feature spatial order and employs it to calculates the possible spatial range of subsequent feature matches, thus filtering out unnecessary feature matches. We also integrate it with epipolar geometry to further improve matching efficiency and accuracy. Since the spatial order of feature points is affected by image rotation, we propose a suitable image alignment method from the fundamental matrix of epipolar geometry to remove the effect of image rotation. To verify the feasibility of the proposed method, we conduct a series of experiments, including a standard benchmark dataset, self-generated simulated images, and real images. The results demonstrate that our proposed method is significantly more efficient and has more accurate feature matching than the traditional method. 	
}

\addKeywords{feature matching, spatial order, guided feature matching, epipolar geometry, image alignment}

\documentclass[\varKeyvals]{\varDocClass}

\usepackage{times}
\usepackage{graphicx}
\usepackage{latexsym}
\usepackage{subfigure}
\usepackage{multirow}
\usepackage{amsmath}
\usepackage{vcell}

\DeclareMathOperator*{\argmax}{argmax} 

\begin{document}
	\insertTopmatter
	
	\makeatletter
	\section{Introduction}\label{section_introduction}
Image feature detection and matching are fundamental techniques in computer vision. Many applications, such as structure from motion, 3D reconstruction, vision-based simultaneous localization and mapping (SLAM), camera pose estimation, image rectification, and image stitching, require a correspondence between image feature points in two images. In the early days of image feature detection, techniques mainly focused on detecting prominent points such as corners in the image, (\textit{e.g.}, Harris's corner detector \cite{Harris88}). A traditional feature matching method is to use a small image patch around the corner to calculate the similarity of two local patches by an image brightness matching technique, such as normalized cross correlation. This similarity is used to determine whether the corner belongs to the same point in a real scene. However, this method is easily affected by changes in illumination, image rotation, zooming, and viewpoint, so it has limitations in practical applications.\\
\indent In contrast to corner detection, some researchers have proposed the so-called blob detectors such as scale-invariant feature transform (SIFT) \cite{Lowe04} and speeded-up robust features (SURF) \cite{Bay08} based on the characteristics of image variation around the feature points. These methods determine feature correspondence by comparing their accompanying well-designed feature descriptors. Due to the effectiveness of these descriptors, these methods are quite discriminative in feature matching, which can effectively overcome the problems of changes in image viewpoint, scaling, rotation, and brightness. Because of the success of these methods, many variations of blob detectors have subsequently been proposed, such as ORB (oriented FAST and rotated BRIEF) \cite{Rublee11}, and many systems were designed based on these techniques, such as Monocular SLAM \cite{Davison07} and ORB SLAM \cite{Mur-Artal15}. \\
\indent Although the dimensions of these feature descriptors are not high, and some are even designed in binary codes (\textit{e.g.}, ORB) to speed up the matching process, they still require considerable time for matching when there are a large number of feature points in an image. Furthermore, the image resolution of current digital cameras is very high and thus tens or hundreds of thousands of feature points may be readily detected in an image. Guided matching is a method that effectively improves the efficiency of image feature matching. It mainly uses some two-view geometry constraints, such as epipolar geometry, to guide or limit the area of feature matching. Epipolar geometry can limit the range of feature matching from a two-dimensional image to a one-dimensional line called an epipolar line, thus significantly eliminating unnecessary matches and speeding up the matching process. Meanwhile, because feature matching is performed only on potential candidates, guided matching can sometimes even improve the accuracy of matching, especially when there are repeating patterns in the image. \\
\indent Epipolar geometry can limit the range of feature matching to a single line, although, in practice, the matching range is usually extended to a band along the epipolar line because of the inevitable error in estimating the epipolar geometry (or, more precisely, the fundamental matrix). Obviously, this increases the number of features to be matched. By combining other models with epipolar geometry to restrict the range of feature matching, the overall performance of feature matching could theoretically be further improved.\\ 
\indent Talker \textit{et al.} proposed using the spatial order of feature points to determine the correctness of a feature match \cite{Talker16, Talker18}. They observed the order of image features in the horizontal direction and found that, if the features were correctly matched, the order was consistent between the two images, while if they were incorrectly matched, the matched features would have order inversion between the matches. Therefore, the correctness of a feature match can be determined to some extent by examining the number of order inversions for this pair of correspondence with others. Based on this concept, Talker \textit{et al.} proposed a method to estimate the number of correct matches from a set of putative feature matches. They also proposed a model to estimate the probability that a match is correct given the number of order inversions in the horizontal direction. \\
\indent  The spatial order of features is another mathematical model that can be used to guide the matching of features to a specific range. By slightly changing the application mode of the spatial order model, we can find that the matched features partition the whole image into many intervals in the horizontal direction. When a new feature is to be matched, its potential correspondences may fall in these intervals. Each interval will thus induce a number of order inversions, which can be used to estimate the probability that the corresponding feature in this interval is a correct match. These probability values can be used to determine which intervals are the likely locations of the corresponding points, which can then be used to guide the matches to a particular range. \\
\begin{figure*}[t]
	\centering
	\includegraphics[height = 4cm]{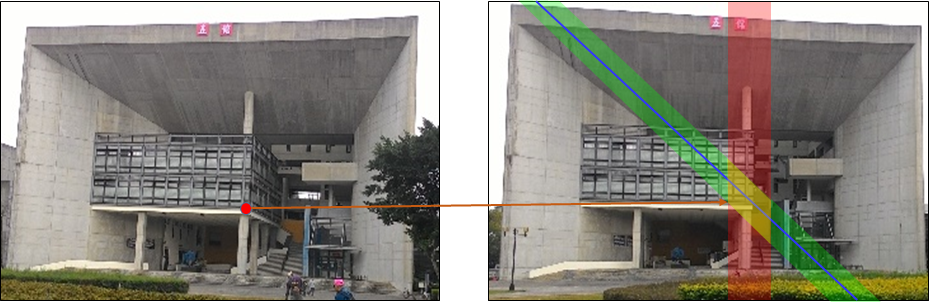}
	\caption{Search region of a feature point. The green area is the search region from the epipolar line (blue line) and the red area is the search region from the spatial order model. The yellow region is the intersection of these two areas and is our final search region for a corresponding point. } \label{figure_search_region}
\end{figure*}
\indent In this paper, we combine the spatial order model with epipolar geometry to achieve more effective image feature matching. The concept of combining these models is illustrated in Fig. \ref{figure_search_region}. The spatial order model leads to a search region for feature matching in the horizontal direction (red area in Fig. \ref{figure_search_region}), while the epipolar geometry generates  a band area along the epipolar line (green area in Fig. \ref{figure_search_region}). The intersection of the two areas (yellow area in Fig. \ref{figure_search_region}) is the final search region for a match. Although the feature spatial order can be used to guide the search range of a match, it is significantly affected by image rotation, limiting its application in practical situations. To solve this problem, we should align the two images to eliminate the rotation component around the optical axis between the two views.\\
\indent Another advantage of combining epipolar geometry with spatial order is that we can calculate the relative rotation of two images from the estimated fundamental matrix. This rotation can then be used to derive a homography matrix, which can be used to align the two images to correct the feature spatial order induced by relative rotation.\\
\indent In this study, we investigate four fundamental matrix-based image alignment approaches and determine a preferred suitable one from a simulated experiment. To effectively utilize the spatial order and epipolar geometry models, we plan a progressive feature matching framework to build a feature search guiding model using the features that were previously matched. This model guides the search region of subsequent feature matches. As the number of matched features increases, the model is progressively updated and  further guides feature matching. This filters out more unnecessary matches so that the overall process of feature matching can be optimized. \\  
\indent The remainder of this paper is organized as follows. We give a brief literature review for feature point detection and matching in the next section. In Section \ref{section_proposed_framework}, we detail our progressive feature matching framework. We first introduce key concepts, including the computation of the spatial order model and how to guide feature matching based on spatial order and epipolar geometry models. We also investigate some image alignment approaches. In Section \ref{section_experimental_results}, we present our experimental results. Finally, we give conclusions in Section \ref{section_conclusion}.

\section{Related Work}
\label{section_related_work}
\subsection{Feature detection}
Feature detection development has a long history. Typically, the techniques of feature detection can be roughly classified into two categories: corner detectors and blob detectors \cite{Herling11}. Corner detectors, such as Harris’s corner detector \cite{Harris88}, quickly detect the corner points in an image. Harris's corner detector employs the auto-correlation matrix of the local area of a point in an image to detect a corner. By analyzing the eigenvalue of this matrix, we can determine whether the point is an edge, a corner, or a point in a smooth area. Shi and Tomasi proposed good features to track (GFTT) \cite{Shi94} and found it improved on Harris's corner detector. Rosten and Drummond proposed the features from accelerated segment test (FAST) feature detector from a different perspective \cite{Rosten06}. This method avoids the eigenvalue analysis and is therefore more efficient to calculate. The advantage of the corner detector approach is its high computational efficiency, but it typically requires robust matching techniques to accommodate issues relating to image rotation, scale variation, and viewpoint change. \\
\indent In contrast to the corner detectors, blob detectors use unique local areas in the image as the detection target. Points in these local areas typically have similar image characteristics. Blob detectors are usually based on Gaussian filtering, which is performed at different image scales to achieve scale-invariant image feature detection. SIFT \cite{Lowe04} is a typical representative of this kind of method. Since SIFT can effectively overcome image brightness variation, image rotation, and scale variation, and has a certain degree of robustness to image viewpoint change, it has been favored by many researchers and is widely used in systems such as image tracking and SLAM \cite{Henry10}. Due to the success of SIFT, many methods were proposed to further improve the detection efficiency and accuracy, with SURF \cite{Bay08} being a representatives example. SURF uses the Hessian matrix after Gaussian convolution to detect feature points. Since it avoids expensive Gaussian filtering, it is more efficient. Although Cheng \textit{et al.} \cite{Cheng07} demonstrated that SURF is superior to SIFT in terms of computational efficiency and robustness, SIFT was found to be comparable to SURF in some subsequent experimental evaluations \cite{Mukherjee15}, and even better than SURF in some cases. \\
\indent Many other detection methods have also been proposed, such as maximally stable extremal regions (MSER) \cite{Matas04}, center surround extremas (CenSurE) \cite{Agrawal08}, binary robust invariant scalable key-points (BRISK) \cite{Leutenegger11}, and ORB \cite{Rublee11}. Most of these methods are based on different criteria. For example, CenSurE has two key criteria: stability and accuracy, with the goal of features being accurately detected and located even when the viewpoint changes. ORB is another feature detection method designed for computational efficiency. It can achieve very efficient feature detection and has high detection accuracy in an experimental evaluation \cite{Mukherjee15}. \\
\indent Recently, researchers have tried to leverage advances in machine learning to design better feature detection systems. Learned invariant feature transform (LIFT) \cite{Yi16} is a complete framework for feature detection and descriptor generation using deep neural networks. It first uses a convolutional neural network (CNN) for feature detection, then performs image alignment for the local area around the feature, and finally generates the feature descriptor, all via deep neural networks. \\
\indent SuperPoint \cite{DeTone18} is another feature detection technique based on machine learning. The authors first synthesized images that contain simple geometric objects such as lines, triangles, quadrilaterals, or cubes. Since the image is synthesized from simple geometric objects, the features can be easily detected. These synthetic images were then used for initial neural network training. By applying what the authors called homographic adaptation, a number of pseudo-ground-truth interest points were detected, which were used to train the neural network with the actual images. The above process was repeated several times to obtain more robust image feature detection than traditional approaches. This method runs on the Titan X GPU and achieves a computing speed of 70 fps for $480\times 640$ images. \\
\indent To conclude, there are many different image feature detection techniques available. Some qualitative and experimental studies \cite{Li08, Mukherjee15} have been conducted to understand the advantages and disadvantages of these methods. While these studies are good references to understand feature detection techniques, they do not cover some of the latest deep-learning-based feature detection techniques.\\

\subsection{Feature matching}
\label{section_feature_matching}
\indent The most intuitive way to match features is to measure the brightness difference of image patches around the features. This concept comes from the idea that there must be similar image brightness or color distribution around the same feature, so the numerical distance of pixel values in the local area around the feature can be used to determine whether the feature comes from the same scene point. However, such image brightness matching is easily affected by illumination variation, viewpoint change, image rotation, and zooming. Thus, it performs poorly in practical applications. Nevertheless, if the image to be matched is from a continuous video, then optical flow techniques such as the Kanade–Lucas–Tomasi (KLT) tracker \cite{Tomasi91} that is commonly used in video analysis, are good choices for feature tracking and matching. However, these sometimes have limitations, including that image motion cannot be too fast and there should not be large image brightness variations. \\
\indent In contrast to image patch matching, feature matching by descriptors is widely used by researchers because it is robust to image brightness variation, viewpoint change, image scaling, and rotation, to some extent. The only drawback of descriptor matching is that the generation of descriptors requires slightly more computations. The generation of descriptors varies according to the chosen technique, \textit{e.g.}, SIFT and SURF each have their own descriptors. For instance, SIFT uses the gradient variations around the feature as its descriptor. Since the SIFT descriptor records the orientation of features, it can effectively overcome the problem of image rotation. \\
\indent Traditional descriptors are generally composed of numbers. To speed up the matching of features, some researchers used binary codes as the descriptor of a feature since the Hamming distance between two binary codes can be calculated quickly by XOR. In addition to dedicated descriptors for specific detection techniques, some researchers have focused only on developing feature descriptors, such as binary robust independent elementary features (BRIEF) \cite{Calonder10} and fast retina key-point (FREAK) \cite{Alahi12}. BRIEF is a binary code descriptor. It can achieve very fast feature matching by taking advantage of the fast computation of binary codes. FREAK is designed based on the human visual system, with the goal of the generated descriptor matching the visual perception of the human eye. These two descriptors do not have their own feature detector, \textit{i.e.}, they can be used with other feature detection techniques. \\
\indent In fact, previously described feature detectors and their associated descriptors can be used in conjunction with each other, sometimes leading to better results. For example, the FAST detector + SIFT descriptor has better experimental results than the original SIFT detector + SIFT descriptor \cite{Mukherjee15}. \\
\indent With the recent achievements of machine learning, some deep-learning-based feature descriptors have been proposed such as binary online learned descriptor (BOLD) \cite{Balntas15}. From about 2015, there has been a growing number of studies proposing deep-learning-based image feature detection and matching. Han \textit{et al.} proposed MatchNet \cite{Han15}, which uses five convolutional layers and three fully connected layers to estimate the similarity of two local image patches. They adopted the Siamese network architecture, which divides the network into two branches to process individual local image patches, where the two branches share network parameters. They also proposed a sampling mechanism to accelerate the training of neural networks by efficiently selecting correct and incorrect matched feature pairs. This method performs better than  traditional methods, such as SIFT, in feature matching. In the same year, Zagoruyko and Komodakis published their study on local image patch matching using CNNs \cite{Zagoruyko15}. Unlike Han \textit{et al.}, they tested and performed experimental evaluations on many network architectures, including 2-channel, Siamese, and multi-resolution architectures. The results confirmed that the 2-channel in conjunction with multi-resolution architecture achieves the most accurate image feature matching. Simo-Serra \textit{et al.} \cite{Simo-Serra15} also used the Siamese network architecture to design their system. However, their architecture did not have fully connected layers. They directly used the output of a CNN as the descriptors of features, and the similarity of features was determined by directly calculating the Euclidean distance between the two descriptor vectors. The advantage of this approach is that some traditional matching techniques, such as approximate nearest neighbors (ANN) \cite{Muja09}, can be applied easily without any modification. \\
\indent DeepBit \cite{Lin16} was proposed by Lin \textit{et al.} They used deep neural networks to generate binary descriptors to improve the efficiency of feature matching by the fast computation of binary codes while maintaining discrimination power. Choy \textit{et al.} \cite{Choy16} proposed a universal correspondence network that effectively detects dense image feature points. This network contains a spatial transformer network \cite{Jaderberg15}, which can normalize the image patches, \textit{i.e.}, it can effectively handle the problems of patch rotation and scaling. Moreover, this method can be used not only to find geometric matching, but also to deal with semantic matching. Although deep-learning-based feature matching generally yields more accurate matching than traditional approaches, it typically requires more computations. \\
\indent Computation time is especially costly when the number of features to be matched is large. Thus, special data structures have been designed to appropriately organize the features to speed up the process of feature matching, with ANN \cite{Muja09} being a common approach. The concept of ANN is that if we do not require perfectly nearest-neighbor search, but can allow a little error, then we can get very fast matching by using some special data structures, such as Kd-Tree, M-tree, or ball tree. Although ANN is several times faster than the brute force approach (\textit{i.e.}, one-to-one feature descriptor matching for the whole dataset), based on our practical experience, it is not as accurate as the brute force approach. \\
\indent In addition to traditional ANN, hashing-based ANN has attracted the attention of some researchers. The hash function can convert the feature descriptors into binary codes and then use the XOR operation of binary codes to significantly speed up the matching. LDAHash \cite{Strecha12} is a feature matching method that uses the hashing framework, but it requires data tagging before training, which is somewhat cumbersome in practical use. Inspired by LDAHash, cascade hashing \cite{Cheng14} is a faster hashing-based image feature matching algorithm. It employs multiple sets of hashing functions to convert features into multiple binary codes for classification to achieve more efficient feature matching. \\
\indent Previous methods mainly use the similarity between feature descriptors for feature classification and hence restrict the matching to specific groups. Another way to speed up feature matching is to guide the matching process in a specific region in the image space. Epipolar geometry, the most famous of these approaches, can limit the matching to a line. In addition to epipolar geometry, there are other methods that can limit feature matching in the image space. For example, Zhu \textit{et al.} \cite{Zhu05} used the established correspondences to partition the image into  a number of triangles, and then used these triangles as constraints to limit subsequent feature matching. However, this method relies heavily on the correctness of the initial matching. If one feature is incorrectly matched, then the created triangular mesh is incorrect, which will lead to incorrect feature matching and error accumulation. Spatial order is another way to guide feature matching in an image space. Since it is not based on epipolar geometry, it can complement epipolar geometry to improve computational efficiency and matching accuracy, which is the main focus of this study. It is worth noting that these spatially constrained methods are not in competition with the aforementioned descriptor-based clustering methods, such as ANN. They can complement each other to achieve more efficient feature matching.

\section{Proposed Framework}
\label{section_proposed_framework}
\subsection{System architecture}
\label{section_system_architecture}
\begin{figure*}
	\centering
	\includegraphics[height = 5cm]{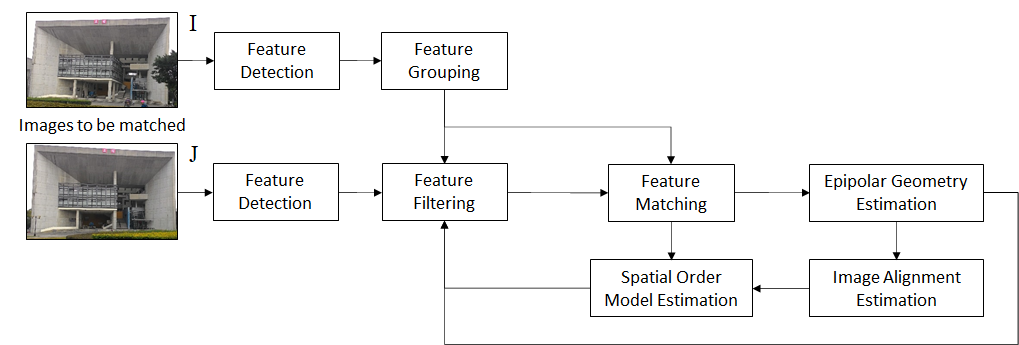}
	\caption{System architecture of the proposed framework.} \label{figure_architecture}
\end{figure*}
The system architecture of the proposed progressive feature matching framework is depicted in Fig. \ref{figure_architecture}. Our system has a feedback structure. In other words, it uses the previously matched features to build the system model (the spatial order and epipolar geometry models) and then the created model is employed to further restrict the searching range of subsequent feature matching. This reduces the required number of matches and hence achieves the goal of efficient image feature matching. There are several modules in this framework, as follows:
\begin{itemize}
	\item {\bf Feature Detection}: Our system does not limit the type of feature detection technique. Any feature detection approach can be employed in our framework.
	\item {\bf Feature Grouping}: To ensure the matched features are distributed on the entire image as uniformly as possible, we partition the image's x-domain into a number of intervals and classify features into these intervals according to their x-coordinates. During the matching process, features are sequentially drawn from these intervals for subsequent feature matching.  
	\item {\bf Feature Matching}: This is ordinary feature matching. For example, this can be achieved by measuring the similarity between feature descriptors. 
	\item {\bf Spatial Order Model Estimation}: This module estimates the required parameters of the spatial order model. We describe the detailed procedures in Section \ref{section_spatial_order}. 
	\item {\bf Epipolar Geometry Estimation}: This module estimates the fundamental matrix using previously matched features by a robust approach such as random sample consensus (RANSAC). This fundamental matrix can be used to estimate the transformation for image alignment and to calculate the corresponding epipolar line in the second image when a new feature is given.
	\item {\bf Feature Filtering}: After obtaining the parameters of the spatial order model and the fundamental matrix, for each feature in the first image, we can calculate the possible region of the corresponding feature on the second image and exclude features outside this region, \textit{i.e.} perform feature filtering. We explain this module in depth in Sections \ref{section_feature_filtering_by_SO} and \ref{section_feature_filtering_by_EG}. 
	\item {\bf Image Alignment Estimation}: This module decomposes the fundamental matrix to find the required transformation matrix for image alignment. The process is detailed in Section \ref{section_image_alignment}.
\end{itemize} 

\indent According to the system architecture in Fig. \ref{figure_architecture}, given two images I and J to be matched, we first detect the features for the two images. We then group the features in image I according to their horizontal positions. Features are then sequentially selected from these groups for matching features in image J. After accumulating a certain number of matched features, we activate the estimation of the spatial order model and the fundamental matrix. After obtaining model information, we use the model to regulate the matching of subsequent features. That is, given a feature in image I, we estimate the possible feature region in image J and exclude features in image J outside this region. In addition, in order to balance the overall computational efficiency of the system, we do not need to estimate the spatial order model and the fundamental matrix for each new feature, but can accumulate a certain number of points before activating the estimation of these two models, thus saving computational resources. In the following, we explain how to estimate model parameters and how to filter the features in image J. \\

\subsection{Spatial order model} \label{section_spatial_order}
\subsubsection{Estimating the number of correct matches}
The spatial order model was first proposed by Talker \textit{et al.} \cite{Talker16, Talker18}. Here, we summarize its main idea and how to estimate it. We follow the notations used by Talker \textit{et al.} \cite{Talker18}. The features in each image must first be sorted using their x-coordinates. Thus, the index of each feature is just its rank in the x-direction. \\
\indent Suppose that the features in the pair of images have been matched using any ordinary feature matching technique. Then, the matching results can be represented by two arrays: $[N]$ and $\sigma$. For example, if $[N]=<1, 2, 3>$ and $\sigma=<3, 2, 1>$, this indicates that the first feature in the first image is matched to the third feature of the second image, the second is matched to the second, and the third is matched to the first. Based on this and some assumptions, the number of correct matches, $N_G$, can be estimated by solving the following quadratic equation \cite{Talker18}:
\begin{equation}\label{Eq:1}
	\frac{1}{6}N^2_G-\bigg(\frac{1}{2}-\frac{N}{3}\bigg)N_G-N(N-1)\bigg(\frac{1}{2}-\hat{K}\bigg)=0,
\end{equation}
where $N$ is the number of matched features and $\hat{K}$ is the normalized Kendall distance, defined as follows: 
\begin{equation}\label{Eq:2}
	\hat{K}([N],\sigma) = \frac{2}{N(N-1)}\sum_{1\leq i\leq N}\sum_{i\leq j\leq N} \eta_\sigma(i, j),
\end{equation}
where $\eta_\sigma(i, j) = \eta_\sigma^r(i, j) + \eta_\sigma^l(i, j)$ and 
\begin{equation}\label{Eq:3}
	\begin{aligned}
		\eta_\sigma^r(i, j) &= 
		\left\{
		\begin{array}{cl}
			1, & \text{if } (i < j)\ \&\ (\sigma(i) > \sigma(j)) \\ 
			0, & \text{otherwise}
		\end{array}\right. \\
		\eta_\sigma^l(i, j) &= 
		\left\{
		\begin{array}{cl}
			1, & \text{if } (i > j)\ \&\ (\sigma(i) < \sigma(j)) \\ 
			0, & \text{otherwise}
		\end{array}\right.
	\end{aligned}
\end{equation}
In fact, the normalized Kendall distance is used to record the normalized number of pairwise order inversions between two sequences. By solving Eq. (\ref{Eq:1}), we can obtain $N_G$, which allows us to calculate the probability that a match is correct as stated in Section \ref{section_matching_probability}. 

\subsubsection{Estimating the overlap region}
Previous estimation of $N_G$ has assumed that the contents of the two images fully overlap. However, this is not always the case and in \cite{Talker18}, a method is proposed to estimate the partially overlapping image region. The overlap region can be defined by a 4-tuple of indices, $\omega^*=(i_L^1, i_H^1, i_L^2, i_H^2)$, where $i_L^1$ and $i_H^1$ respectively denote the lowest and highest indices of correct matches in the first image, and $i_L^2$ and $i_H^2$ are respectively the lowest and highest for the second image. With this notation, Eq. (\ref{Eq:1}) can be transformed to the following equation as $N$, $N_G$, and $\hat{K}$ are now functions of $\omega^*$:    
\begin{equation}
	\begin{split}
		\frac{1}{6}N_G^2(\omega^*)
		- \bigg(\frac{1}{2}-\frac{N(\omega^*)}{3}\bigg) N_G(\omega^*) \\
		- N(\omega^*)(N(\omega^*)-1)
		\bigg(\frac{1}{2}-\hat{K}(\omega^*)\bigg) = 0.
	\end{split}
\end{equation}
Talker \textit{et al.} proved that the overlap region can be estimated by finding the maximum of $N_G(\omega^*)$ \cite{Talker18}, \textit{i.e.}, 
\begin{equation}
	\omega^* = \argmax\limits_{\omega} \thinspace N_G(\omega).
\end{equation}
In our progressive feature matching framework, if an incoming feature is outside the overlap region, then it is immediately discarded without further feature matching.

\subsubsection{Estimating the matching probability} \label{section_matching_probability}
\begin{figure}
	\centering
	\includegraphics[height=4cm]{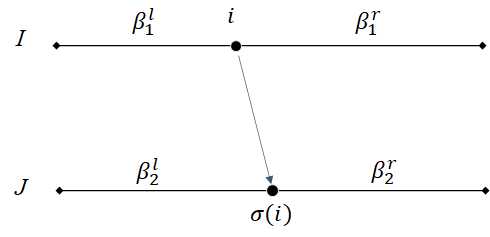}
	\caption{The number of incorrect matches in the left and right parts of the match $(i, \sigma(i))$.} \label{figure_number_incorrect_matches}
\end{figure}
To estimate the probability that a match is correct, we first compute $H_\sigma(i)$, the number of inversions in which a match $(i,\sigma(i))$ participates. In particular, $H_\sigma(i)$ can be expressed as the sum of two terms, \textit{i.e.}, $H_\sigma(i) = H_\sigma^l(i) + H_\sigma^r(i)$, where $H_\sigma^l(i) = \sum_{j<i}\eta_\sigma^l(i, j)$ is the number of inversions from the left of $i$ to the right of $\sigma(i)$ and  $H_\sigma^r(i) = \sum_{j>i}\eta_\sigma^r(i, j)$ is the number of inversions from the right of $i$ to the left of $\sigma(i)$. With $H_\sigma^l(i)$ and $H_\sigma^r(i)$, the probability that the $i$th match is correct is estimated as follows:
\begin{equation}
	\begin{aligned}
		&P(i \in G \mid H_{\sigma}^l(i), H_{\sigma}^r(i))
		= \\
		&\frac{
			P_{H_{\sigma}^l, H_{\sigma}^r, i \in G} P(i \in G)
		}{
			P_{H_{\sigma}^l, H_{\sigma}^r, i \in G} P(i \in G) + P_{H_{\sigma}^l, H_{\sigma}^r, i \notin G} P(i \notin G)
		}.
	\end{aligned}
\end{equation}
where $G$ denotes the set of correct matches and $ P_{H_{\sigma}^l H_{\sigma}^r, i \in G} = P(H_{\sigma}^l(i), H_{\sigma}^r(i)|i \in G)$, $P_{H_{\sigma}^l, H_{\sigma}^r, i \notin G} = P(H_{\sigma}^l(i), H_{\sigma}^r(i)|i \notin G)$.
The probability $P(H_{\sigma}^l(i), H_{\sigma}^r(i)|i \in G)$ is computed by 
\begin{equation} \label{Eq:7}
	\begin{aligned}
	& P(H_{\sigma}^l(i), H_{\sigma}^r(i)|i \in G)=\\
	& \sum_{\beta \in S_{\beta}}P(H_{\sigma}^l(i), H_{\sigma}^r(i)|i \in G, \beta)P(\beta),
	\end{aligned}
\end{equation}
where $\beta =(\beta_1^l, \beta_1^r, \beta_2^l, \beta_2^r)$ indicates the number of incorrect matches in the left and right parts of the match $(i, \sigma(i))$, as shown in Fig. \ref{figure_number_incorrect_matches}, and $S_\beta$ is the set of all possible values of $\beta$. The probability $P(\beta)$ can be estimated by multiplying two hypergeometric probabilities as follows:
\begin{equation}
	P(\beta) = \mathcal{H}(i-1, \beta^{l}_{1};N,N_B)\mathcal{H}(\sigma(i)-1,\beta^{l}_{2};N,N_B),
	\label{Eq:8}
\end{equation}
where $\mathcal{H}(n, k; N, K)={K \choose k}{N-K \choose n-k}/{N \choose n}$ is the probability density function of a hypergeometric distribution, which describes the probability of $k$ successes in $n$ draws without replacement from a population of size $N$ that contains exactly $K$ successes. Eq. (\ref{Eq:8}) comes from the fact that we need to select $\beta^l_1$ and $\beta^l_2$ points from the left of $i$ and $\sigma(i)$ with population size $N$ that contains $N_{B}$ fails ($N_{B}$ is the number of incorrect matches, \textit{i.e.}, $N_{B} = N - N_{G}$). Thus, their probabilities clearly follow the hypergeometric distribution. \\
\indent The probability $P(H_{\sigma}^l(i), H_{\sigma}^r(i)|i \in G, \beta)$ is formulated as follows:
\begin{equation}
	\begin{aligned}	
	& P(H_{\sigma}^l(i), H_{\sigma}^r(i)|i \in G, \beta) = \\
	& \mathcal{H}(\beta^{l}_{1}, H_{\sigma}^l(i);N_B,N_B-\beta^{l}_{2})\mathcal{H}(\beta^{l}_{2}, H_{\sigma}^r(i);N_B,N_B-\beta^{l}_{1}).
	\end{aligned}
\end{equation}
The formulation is similar to $P(\beta)$, but now we have $H_{\sigma}^l(i)$ inversions on the left of $i$ and $H_{\sigma}^r(i)$ inversions on the left of $\sigma(i)$.  Thus, $P(H_{\sigma}^l(i), H_{\sigma}^r(i)|i \in G, \beta)$ is the multiplication of the two hypergeometric probabilities. Finally, we need to calculate $P(H_{\sigma}^l(i), H_{\sigma}^r(i)|i \notin G)$. Its computation is similar to Eq. (\ref{Eq:7}) with slight modifications. \\
\indent To improve computational efficiency, some of the above computations are simplified. For example, the hypergeometric distribution can be approximated by a Gaussian function and only a subset of $S_{\beta}$ is used in the computation \cite{Talker18}. Moreover, the probability $P(H_{\sigma}^l(i), H_{\sigma}^r(i)|i \notin G)$ is modeled as a uniform distribution \cite{Talker18}. In our framework, the estimated probability can be used to select the intervals a feature to be matched should fall, thus achieving the goal of filtering out undesired feature matches and improving the matching efficiency.   

\subsection{Feature filtering based on spatial order} \label{section_feature_filtering_by_SO}
\begin{figure}
	\centering
	\includegraphics[height=6cm]{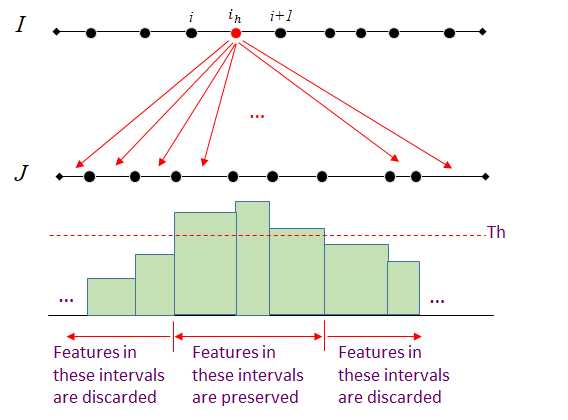}
	\caption{A feature may map to many intervals with different numbers of order inversions. However, only the intervals with matching probability higher than a threshold (Th) are considered in the feature matching. } \label{figure_search_interval}
\end{figure}
After establishing the spatial order model, we can employ it to filter out undesirable feature matches. Suppose we have an incoming feature located between features $i$ and $i+1$ as shown in Fig. \ref{figure_search_interval}. This feature may map to $N+1$ intervals in the second image with each interval associated with a number of order inversions, $H_k$. By temporarily setting the matched feature indices in the $k$th interval as $(i_h, \sigma(i_h)=j_k)$ and letting $i_h = i+0.5$ and $j_k = k+0.5$, we can then calculate $H_k$ as follows:
\begin{equation}
	H_k= \sum_{1\leq j \leq N}\eta_{\sigma}(i_h, j).
\end{equation}
With $H_k$, we can estimate the probability that a correct match occurred in this interval. Typically, the interval with the least number of order inversions can produce the highest probability that a match is correct. With the increased number of order inversions, the probability decreases. Hence, by thresholding the probabilities, we can find a set of intervals that determines the index range from which the features in the second image should be selected for subsequent feature matches. In other words, this selection mechanism allows us to filter out many undesired features in the sense that their matching probabilities are below a predefined threshold value.    

\subsection{Feature filtering based on epipolar geometry} \label{section_feature_filtering_by_EG}
Using epipolar geometry to limit search range is a fundamental technique commonly used for image feature matching. The basic concept is that if a feature in space remains motionless when the images are taken, then the corresponding points in the two images satisfy a relationship called epipolar geometry, which can be expressed as:
\begin{equation}
	{\bf x'}^T{\bf Fx} = 0,
\end{equation}
where ${\bf x}$ and ${\bf x'}$ are the corresponding points in the two images, and ${\bf F}$ is the so-called fundamental matrix, which can be estimated from the corresponding points by a robust method, such as RANSAC. Once ${\bf F}$ is obtained, given a feature ${\bf x}$ on the first image, its epipolar line ${\bf l}$ on the second image can be calculated as ${\bf l}={\bf Fx}$. Theoretically, the corresponding point ${\bf x'}$ of ${\bf x}$ should fall on ${\bf l}$. However, there may be estimation errors so we allow some flexibility. In this study, if the distance between a point ${\bf x'}$ and the epipolar line ${\bf l}$ is larger than a threshold value, then this point is filtered out during the matching process.

\subsection{Image alignment} \label{section_image_alignment}
Since the spatial order of feature points will be altered by image rotation, we need to remove the relative rotation (specifically, the rotation around the optical-axis or Z-axis) between the two images to make the spatial order model work properly. We call this step {\it image alignment}. \\
\indent Suppose we have two images taken for the same scene. Then, according to the camera pinhole model, the points projected on these two images satisfy the following equations:
\begin{equation}
	{\bf m}_1 = {\bf K}_1[{\bf I}|{\bf 0}]{\bf M}
\end{equation}
and
\begin{equation}
	{\bf m}_2 = {\bf K}_2[{\bf R}|-{\bf R}{\bf t}]{\bf M}, 
\end{equation}
where ${\bf M}$ is a scene point in three-dimensional (3D) space, ${\bf m}_1$ and ${\bf m}_2$ are the projected points on the two images, ${\bf K}_1$ and ${\bf K}_2$ are the associated camera calibration matrices of the two images, and ${\bf R}$ and {\bf t} are the relative orientation and displacement of the cameras when the two images were taken. \\
\indent To align the two images, we first determine the relative rotation matrix ${\bf R}$ of the two images. In this study we employ the approach proposed by Hartley \cite{Hartley04}. Specifically, if the camera calibration matrices, ${\bf K}_1$ and ${\bf K}_2$, are known, we can calculate the essential matrix ${\bf E}={\bf K}_2^T {\bf F}{\bf K}_1$ from the fundamental matrix ${\bf F}$. We then perform singular value decomposition (SVD) on ${\bf E}$ to obtain ${\bf E}={\bf UDV}^T$. Based on Hartley's argument, there are two rotation matrices that meet this camera configuration, \textit{i.e.}, 
\begin{equation}
	{\bf R}={\bf UWV}^T \quad \text{or} \quad {\bf R}={\bf UW}^T{\bf V}^T,  
\end{equation} 
where 
\begin{equation*}
{\bf W}= \begin{bmatrix} 0 & -1 & 0 \\ 1 & 0 & 0 \\ 0 & 0 & 1 \\ \end{bmatrix}.	
\end{equation*} 
\indent The translation vector ${\bf t}$, which conforms to the projection equation, is the third column ${\bf u}_3$ of the matrix ${\bf U}$. In practice, ${\bf t}=-{\bf u}_3$ also conforms to the projection equation, and by the obtained two ${\bf R}$s, there are four combinations for ${\bf R}$ and ${\bf t}$. These four combinations represent four camera configurations of the two views, but only one of them can make the reconstructed 3D points fall in front of the two views. Thus, the correct rotation matrix ${\bf R}$ can be found by examining the relative positions of the reconstructed 3D points and the two views \cite{Hartley04}. \\
\indent However, the above analysis assumes that we have acquired ${\bf K}_1$ and ${\bf K}_2$, but in reality, ${\bf K}_1$ and ${\bf K}_2$ are not necessarily known. In this study, we assume that the principal point in the camera calibration matrix is at the center of the image, the skew factor is $0$, and the aspect ratio is $1$. Although these are only assumptions, they work well in real situations. In the camera calibration matrix, only the focal length is unknown. Here, we use Bougnoux's method \cite{Bougnoux98} to estimate the focal length of the first view as follows:
\begin{equation}
	f_1 = \sqrt{-\frac{{\bf p}_2^T[{\bf e}_2]_\times \tilde{{\bf I}}{\bf F}{\bf p}_1{\bf p}_1^T{\bf F}^T{\bf p}_2}
		{{\bf p}_2^T[{\bf e}_2]_\times \tilde{{\bf I}}{\bf F}\tilde{{\bf I}}{\bf F}^T{\bf p}_2}},
\end{equation}
where ${\bf F}$ is the fundamental matrix, ${\bf p}_1$ and ${\bf p}_2$ respectively represent the principal points of the first and second views, ${\bf e}_2$ is the epipole on the second view (\textit{i.e.}, the projection of the optical center of the first view on the second view), $[{\bf e}_2 ]_{\times}$ is the skew-symmetric matrix generated from ${\bf e}_2$, and the matrix $\tilde{\bf I} = \text{diag}(1,1,0)$. The above equation can be used to obtain the focal length of the first view. However, since the roles of the two views are interchangeable, the focal length $f_2$ corresponding to the second view can be estimated by replacing ${\bf F}$ with ${\bf F}^T$, ${\bf p}_1$ with ${\bf p}_2$, and ${\bf e}_2$ with ${\bf e}_1$ in the above equation. \\
\indent In practice, the calculation of focal length is not very stable, and sometimes it is even impossible to find a reasonable focal length because the value in the square root is negative. Fortunately, in our experiments, we found that the values of the focal length are not very sensitive to the estimation of the rotation matrix. In other words, when the focal length cannot be found, a relatively reasonable rotation matrix ${\bf R}$ can still be obtained even with a guessed focal length. According to \cite{Fusiello11}, the focal length value should be between $(w+h)/3$ and $3(w+h)$, where $w$ and $h$ are the width and height of the image, respectively. In this study, if the estimated focal length is not in this range, we directly set the focal length to $w+h$. \\
\begin{figure*}
	\centering
	\includegraphics[height=4.0cm]{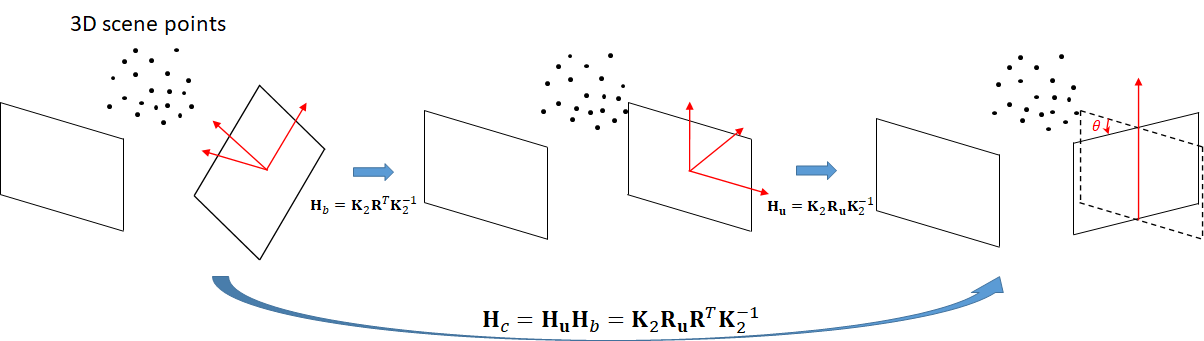}
	\caption{Illustration of our second and third image alignment approaches.} \label{figure_rectification_concept}
\end{figure*}
\indent There are several ways to align the two images. In this study, we try the following four approaches and analyze their performance using a simulated experiment:
\begin{enumerate}
	\item{\bf Image Alignment by ${\bf R}_Z$}: To align the two images, we can remove the rotational component around the view direction (optical axis or Z-axis) of the second image. For this purpose, we decompose the matrix ${\bf R}$ by Euler angles, \textit{i.e.}, ${\bf R}={\bf R}_Z {\bf R}_Y {\bf R}_X$, where ${\bf R}_X$, ${\bf R}_Y$, and ${\bf R}_Z$ are the rotation matrices around the X-axis, Y-axis, and Z-axis, respectively. We then create a transformation ${\bf H}_a = {\bf K}_2{\bf R}_Z^T{\bf K}_2^{-1}$ and apply it to the second view, \textit{i.e.}, 
	\begin{equation}
		{\bf H}_a{\bf m}_2 = {\bf K}_2[{\bf R}_Y{\bf R}_X|-{\bf R}_Y{\bf R}_X{\bf t}]{\bf M}.
	\end{equation}
	From this equation, we can see that the Z-axis rotation on the second view has been removed. This is our first image alignment approach.
	\item{\bf Image Alignment by ${\bf R}$}: Since the relative rotation between the two images is ${\bf R}$, we can directly set ${\bf H}_b={\bf K}_2 {\bf R}^T {\bf K}_2^{-1}$ to represent the transformation on the second view, obtaining
	\begin{equation}
		{\bf H}_b{\bf m}_2 = {\bf K}_2[{\bf I}|-{\bf t}]{\bf M}.
		\label{Eq:17}
	\end{equation}
	Compared with the imaging equation of the first view, we can find that the relative rotation between the two views has been eliminated. The basic concept of this approach is illustrated in the middle of Fig. \ref{figure_rectification_concept}.	
	\item{\bf Image Alignment by ${\bf R}$ and ${\bf R_u}$}: Although the second approach can make the two views face in the same direction, it may cause the second view to deviate from the original viewing direction, seriously distorting the results, especially when the scene falls outside the visible range of the view. Therefore, in this approach, we try to turn back the viewing direction while preserving the transformed Z-axis rotation of the second view, so that it has a similar Z-axis rotation as the first view. To turn the viewing direction back, we inspect Eq. (\ref{Eq:17}). We see that, after the transformation of ${\bf H}_b$, the corresponding rotation matrix changes from ${\bf R}$ to ${\bf I}$. Note that the third column of the rotation matrix represents the viewing direction of the camera. Thus, after the transformation of ${\bf H}_b$, the viewing direction changes from ${\bf r}_3$, the third column of ${\bf R}$, to $[0\quad 0\quad 1]^T$. Let $\theta$ denote the angle between ${\bf r}_3$ and $[0 \quad 0 \quad 1]^T$ and let ${\bf R_u} = {\bf R}\{{\bf u}, \theta\}$, where ${\bf u} = {\bf r}_3 \times [0\quad 0\quad 1]^T$ and ${\bf R}\{{\bf u}, \theta\}$ denotes the rotation matrix around axis ${\bf u}$ with angle $\theta$. Then, we can define the transformation of our third image alignment approach by ${\bf H}_c = {\bf K}_2{\bf R_u}{\bf R}^T{\bf K}_2^{-1}$ and the resulting transformation on the second view becomes:
	\begin{equation}
		{\bf H}_c{\bf m}_2 = {\bf K}_2[{\bf R_u}|-{\bf R_u}{\bf t}]{\bf M}.
	\end{equation}
	The concept of this approach is illustrated in the right of Fig. \ref{figure_rectification_concept}.
	\item{\bf Image Alignment by SVD}: In addition to the above approaches, a stereo rectification method can be used to align the images \cite{Wu18}. This approach is mainly achieved by using the SVD of the essential matrix and it requires the two views (${\bf H}_L$ and ${\bf H}_R$) to be transformed simultaneously. The transformation equations are defined as follows:
	\begin{equation}
		{\bf H}_L = {\bf K}{\bf R}_X^T{\bf R}_L{\bf K}^{-1}
	\end{equation}
	and
	\begin{equation}
		{\bf H}_R = {\bf K}{\bf R}_X^T{\bf R}_R{\bf K}^{-1},
	\end{equation}
	where ${\bf K} = ({\bf K}_1+{\bf K}_2)/2$, ${\bf R}_L = {\bf R}_G {\bf U}^T{\bf R}$, ${\bf R}_R = {\bf R}_G {\bf U'}^T{\bf R}^T$, ${\bf U'} = {\bf R}^T{\bf U}\text{diag}(1, \sigma, 1)$ and the SVD of the essential matrix is ${\bf E} = {\bf U}\text{diag}(1, \sigma, 0){\bf V}^T$. The definition of ${\bf R}_G$ is 
	\begin{equation*}
		{\bf R}_G=
		\begin{bmatrix} 0 & 0 & 1 \\ 0 & 1 & 0 \\ -1 & 0 & 0 \\ \end{bmatrix},
	\end{equation*}
	and ${\bf R}_X$ is the X-axis rotation matrix from the Euler decomposition of ${\bf R}_L$, \textit{i.e.}, ${\bf R}_L = {\bf R}_X {\bf R}_Y {\bf R}_Z$.
\end{enumerate}
\begin{figure*}
	\centering
	\includegraphics[height=7cm]{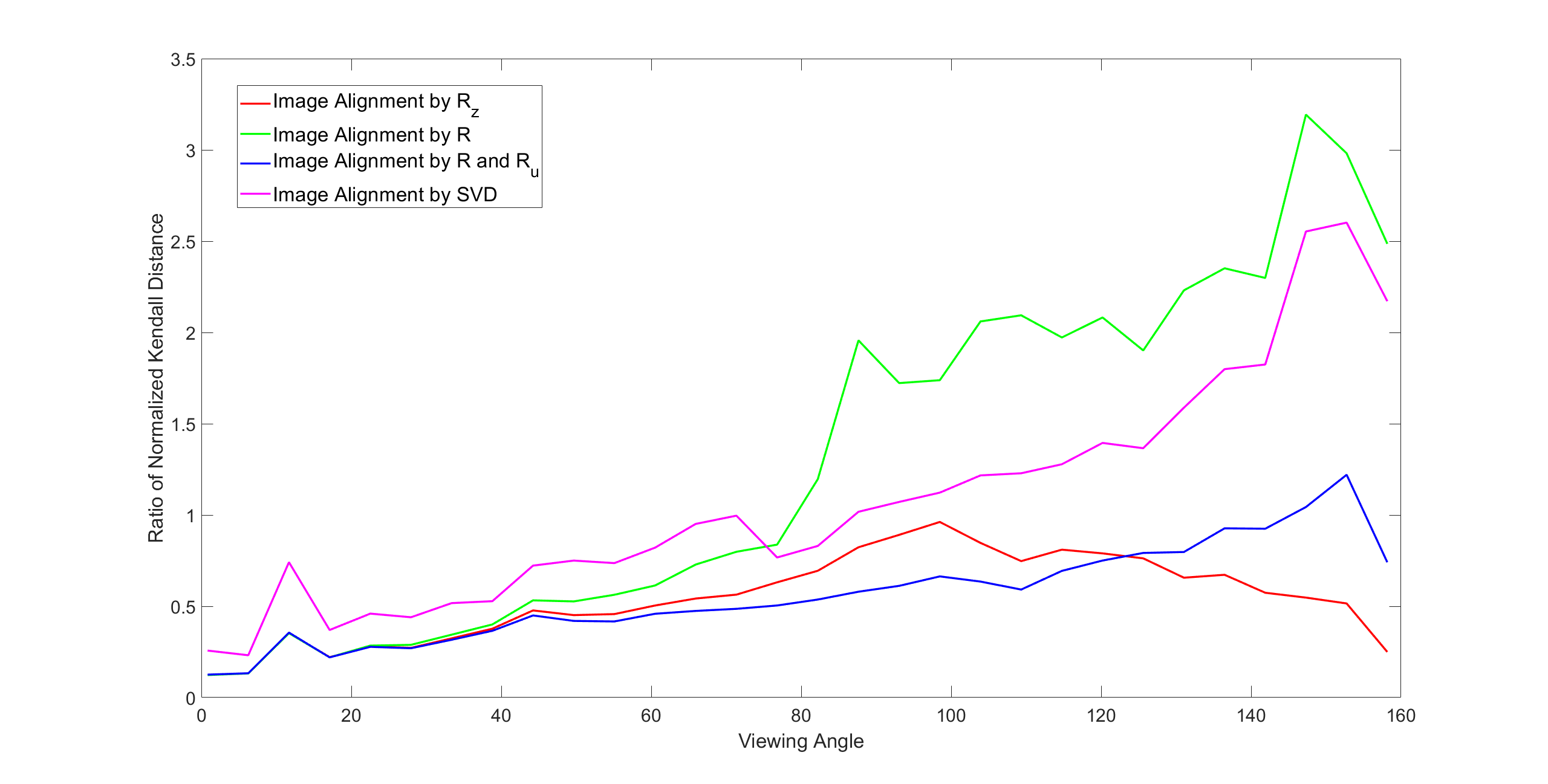}
	\caption{Ratio of normalized Kendall distance vs. viewing angle for four image alignment approaches.} \label{figure_rectification}
\end{figure*}

\indent To verify the feasibility of the four approaches mentioned above, we conducted a simulation experiment in which we generated some random 3D points in space, and then used a virtual camera to generate two views of these 3D points. The camera was placed on the surface of a sphere, facing the center of the sphere (\textit{i.e.}, toward the 3D points). The position of the camera on the sphere, the distance from the camera to the center of the sphere, and the rotation of the camera around the view axis were all randomly generated. In addition, a Gaussian noise with a standard deviation of 1 pixel was added to the image correspondences to simulate the real situation. Typically, image rotation will increase the number of order inversions. Thus, if the rotational component between the two images are removed, the number of order inversions will decrease. Therefore, we employ the normalized Kendall distance (see Eq. (\ref{Eq:2})) of these correspondences to evaluate the effect of image alignment. We calculate the ratio of normalized Kendall distance before and after the image alignment, and if the ratio is less than 1, the image alignment can effectively remove the rotational component between the two views.  \\
\indent We generated a total of 3000 pairs of views and the experimental results are shown in Fig. \ref{figure_rectification}. The horizontal axis of Fig. \ref{figure_rectification} is the angle between the view direction of the two views, and the vertical axis is the ratio of the normalized Kendall distance. As can be seen, the third approach (\textit{i.e.}, image alignment by ${\bf R}$ and ${\bf R_u}$) yields the best results, with the ratio of the normalized Kendall distance being lower than that of the other methods in most cases (especially for low viewing angles, which are the cases in most practical situations). Furthermore, the values are below 1 in most cases. This indicates that the third image alignment approach can perform well, and the number of order inversions is low. Therefore, in the subsequent experiments, we use this approach for image alignment.\\

\section{Experimental Results} \label{section_experimental_results}
To evaluate the performance of our progressive feature matching framework, we conducted a series of experiments. The datasets used in these experiments consisted of a commonly used benchmark image set, a set of self-generated simulated images, and some real images. Our experimental environment was a PC with an AMD Ryzen 9 3900 12-Core processor with a 3.79 GHz CPU and 64 GB RAM. Our system does not limit the used feature detection and matching techniques, but in our subsequent experiments, we use SURF for feature detection and its accompanying descriptors for feature matching. In our experiments, the spatial order and epipolar geometry models are updated after accumulating 200 new correspondences, and a total of three updates are performed. When using the spatial order model for feature filtering, if the probability for a correct match of an interval is less than 0.01, then the features located in this interval are filtered out. When using epipolar geometry to filter features, if the distance between a feature and the epipolar line is greater than five pixels, then it is filtered out. \\
\indent In the following experiments, we used the method of brute force matching as the basis of comparison with the aim of seeing how much our system can improve on the most basic feature matching strategy. \\
\indent In fact, our framework is not in competition with other descriptor-based matching techniques such as ANN, but can be used in combination with them to achieve better matching efficiency. We give our detailed experimental results and discuss the three datasets used.\\
\subsection{Oxford dataset}
The Oxford image dataset \cite{Mikolajczyk05} is a standard test set commonly used in the field of image feature matching. It can be used to evaluate the performance of different feature detection and matching techniques. The images in this dataset contain various geometric and photometric transformations, including blurring, viewpoint change, scaling, and rotation. In addition to images, the dataset also provides homography matrices between images, so that any point in one image can be used to find its correspondence in another image through these homography matrices. This allows us to evaluate whether a correspondence between two images is correct or not. Since there is calculation error in the process of image feature detection and localization, in this study, as long as the distance between the identified correspondence and the correct point calculated from homography was less than three pixels, we considered this a correct match. \\
\begin{figure*}
	\centering
	\begin{tabular}[t]{ccc}
		\subfigure[1000$\times$700, image blur]{\includegraphics[height=3cm]{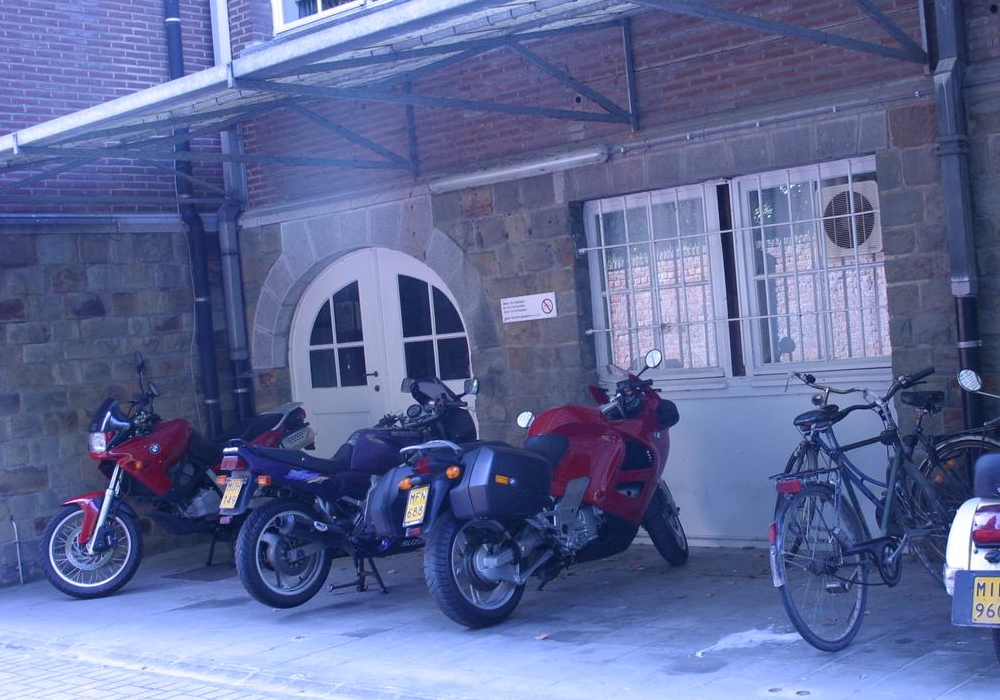}} \hspace{0.2cm}
		\subfigure[1000$\times$700, viewpoint change]{\includegraphics[height=3cm]{wall.png}} \hspace{0.2cm}
		\subfigure[850$\times$680, scale \& rotation]{\includegraphics[height=3cm]{boat.png}} \\
		\subfigure[900$\times$600, illumination change]{\includegraphics[height=3cm]{leuven.png}} \hspace{0.2cm}
		\subfigure[800$\times$640, jpeg compression]{\includegraphics[height=3cm]{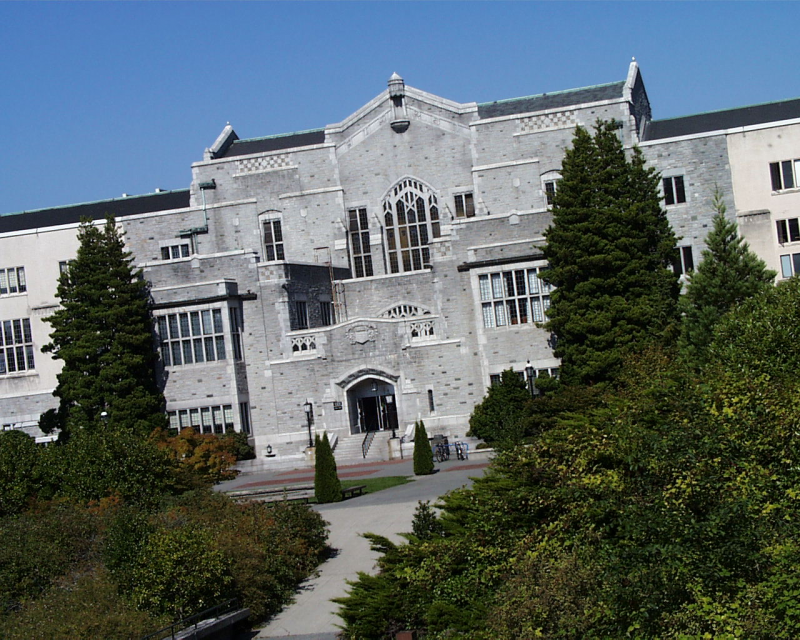}} \hspace{0.2cm}
		\subfigure[800$\times$640, viewpoint change]{\includegraphics[height=3cm]{graf.png}}						
	\end{tabular}
	\caption{Test images from the Oxford dataset.}
	\label{figure_Oxford_images}
\end{figure*}
\indent Each image set in the Oxford image dataset contains at least six images, named imgN, where N represents the image index. Figure \ref{figure_Oxford_images} shows the six test images in the dataset. Some images in the Oxford dataset have very severe distortion and variations, which can be used to verify the feasibility of feature detection and descriptor generation algorithms. However, the proposed framework is not designed to test whether the feature can be detected robustly or whether the feature descriptors can resist large image variations. Instead, our framework is designed to focus on how to guide feature matching in an image space to improve computational efficiency and matching accuracy. Therefore, in this study, we did not test images with severe deformation in the Oxford image dataset, but only used img1-img3 for testing. Nonetheless, by employing more robust feature detection and matching algorithms in our framework, it would be possible to deal with severely deformed images.\\
\begin{table*}
	\footnotesize
	\centering
	\caption{Number of matches for the images in Fig. \ref{figure_Oxford_images}.}
	\label{table:matches}
	\begin{tabular}{p{1.2cm}p{1.7cm}cccc} 
		\hline
		Image & Image Pair & BF & SO & SO+EPI & SO+EPI+IA \\ \hline
		\multirow{2}{1.2cm}{Fig. \ref{figure_Oxford_images}(a)} & img1 vs. img2 & 14,610,528 & 1,628,751 & 456,829 & 457,039\\ 
		& img1 vs. img3 & 12,920,400 & 1,881,958 & 462,219 & 462,601 \\ \hline
		\multirow{2}{1.2cm}{Fig. \ref{figure_Oxford_images}(b)} & img1 vs. img2 & 55,041,906 & 7,746,167 & 1,147,349 & 1,154,311\\ 
		& img1 vs. img3 & 53,536,529 & 16,763,971 & 1,555,961 & 1,537,344 \\ \hline
		\multirow{2}{1.2cm}{Fig. \ref{figure_Oxford_images}(c)} & img1 vs. img2 & 31,215,078 & 9,591,031 & 1,475,852 & 1,361,581\\ 
		& img1 vs. img3 & 25,362,588 & 8,137,117 & 1,602,778 & 1,509,089 \\ \hline
		\multirow{2}{1.2cm}{Fig. \ref{figure_Oxford_images}(d)} & img1 vs. img2 & 13,765,075 & 1,848,385 & 448,676 & 447,671\\ 
		& img1 vs. img3 & 11,848,012 & 1,962,436 & 430,578 & 430,578 \\ \hline
		\multirow{2}{1.2cm}{Fig. \ref{figure_Oxford_images}(e)} & img1 vs. img2 & 23,540,460 & 1,353,087 & 557,546 & 557,546\\ 
		& img1 vs. img3 & 22,919,430 & 1,326,456 & 538,262 & 537,556 \\ \hline
		\multirow{2}{1.2cm}{Fig. \ref{figure_Oxford_images}(f)} & img1 vs. img2 & 17,253,507 & 5,893,223 & 913,595 & 895,077\\ 
		& img1 vs. img3 & 18,953,805 & 8,569,099 & 2,968,126 & 2,951,364 \\ \hline
	\end{tabular}
	\label{table_matches} 
\end{table*}
\indent Table \ref{table_matches} lists the required number of feature matches for our framework and the brute force approach, where BF stands for brute force, SO means only the spatial order model is used for feature filtering, SO+EPI indicates that both the spatial order and epipolar geometry models are used for feature filtering, and SO+EPI+IA indicates use of spatial order, epipolar geometry, and image alignment. \\
\indent Since the brute force approach is a one to one match for all features in both images, the number of matches required is the product of the number of detected features in the two images. By contrast, the other approaches require far fewer feature matches than the brute force method because many features are filtered out by the spatial order or epipolar geometry models during the matching process. From Table \ref{table_matches}, we find that by using only the spatial order model, the number of matches can be reduced to between 5\% and 45\% of that of the brute force approach. If we combine spatial order and epipolar geometry, then the number of matches can be reduced to only 2\% to 15\% of the brute force approach. This data demonstrates that our progressive feature matching framework based on spatial order and epipolar geometry is effective in reducing the number of matches. Finally, the number of matches of SO+EPI+IA is comparable with SO+EPI, indicating that our image alignment does not increase the number of feature matches, and in some cases, it is even slightly decreased. \\
\begin{table*}
	\footnotesize
	\centering
	\caption{Number of detected true correspondences for the images in Fig. \ref{figure_Oxford_images}.}
	\label{table:inliers}
	\begin{tabular}{p{1.2cm}p{1.7cm}cccc} 
		\hline
		Image & Image Pair & BF & SO & SO+EPI & SO+EPI+IA \\ \hline
		\multirow{2}{1.2cm}{Fig. \ref{figure_Oxford_images}(a)} & img1 vs. img2 & 1,806 & 1,828 & 1,844 & 1,849\\ 
		& img1 vs. img3 & 1,319 & 1,319 & 1,324 & 1,317 \\ \hline
		\multirow{2}{1.2cm}{Fig. \ref{figure_Oxford_images}(b)} & img1 vs. img2 & 2,600 & 2,589 & 2,592 & 2,585\\ 
		& img1 vs. img3 & 1,660 & 1,611 & 1,595 & 1,625 \\ \hline
		\multirow{2}{1.2cm}{Fig. \ref{figure_Oxford_images}(c)} & img1 vs. img2 & 958 & 574 & 556 & 971\\ 
		& img1 vs. img3 & 502 & 253 & 260 & 518 \\ \hline
		\multirow{2}{1.2cm}{Fig. \ref{figure_Oxford_images}(d)} & img1 vs. img2 & 1,984 & 2,028 & 2,056 & 2,056\\ 
		& img1 vs. img3 & 1,449 & 1,498 & 1,505 & 1,505 \\ \hline
		\multirow{2}{1.2cm}{Fig. \ref{figure_Oxford_images}(e)} & img1 vs. img2 & 3,113 & 3,148 & 3,151 & 3,151\\ 
		& img1 vs. img3 & 2,458 & 2,490 & 2,497 & 2,496 \\ \hline
		\multirow{2}{1.2cm}{Fig. \ref{figure_Oxford_images}(f)} & img1 vs. img2 & 711 & 384 & 450 & 801\\ 
		& img1 vs. img3 & 208 & 163 & 214 & 275 \\ \hline
	\end{tabular}
	\label{table_true_matches} 
\end{table*}
\begin{figure*}
	\centering
	\begin{tabular}{@{}c@{}lcr@{}}
		\savecellbox{\rule{0pt}{4.698cm}}
		& \savecellbox{\subfigure[]{\includegraphics[height=3cm]{boat.png}}}  \hspace{0.5cm}
		& \savecellbox{\subfigure[]{\includegraphics[height=3cm]{boat3.png}}}  \hspace{0.5cm}
		& \savecellbox{\subfigure[]{\includegraphics[height=4.698cm]{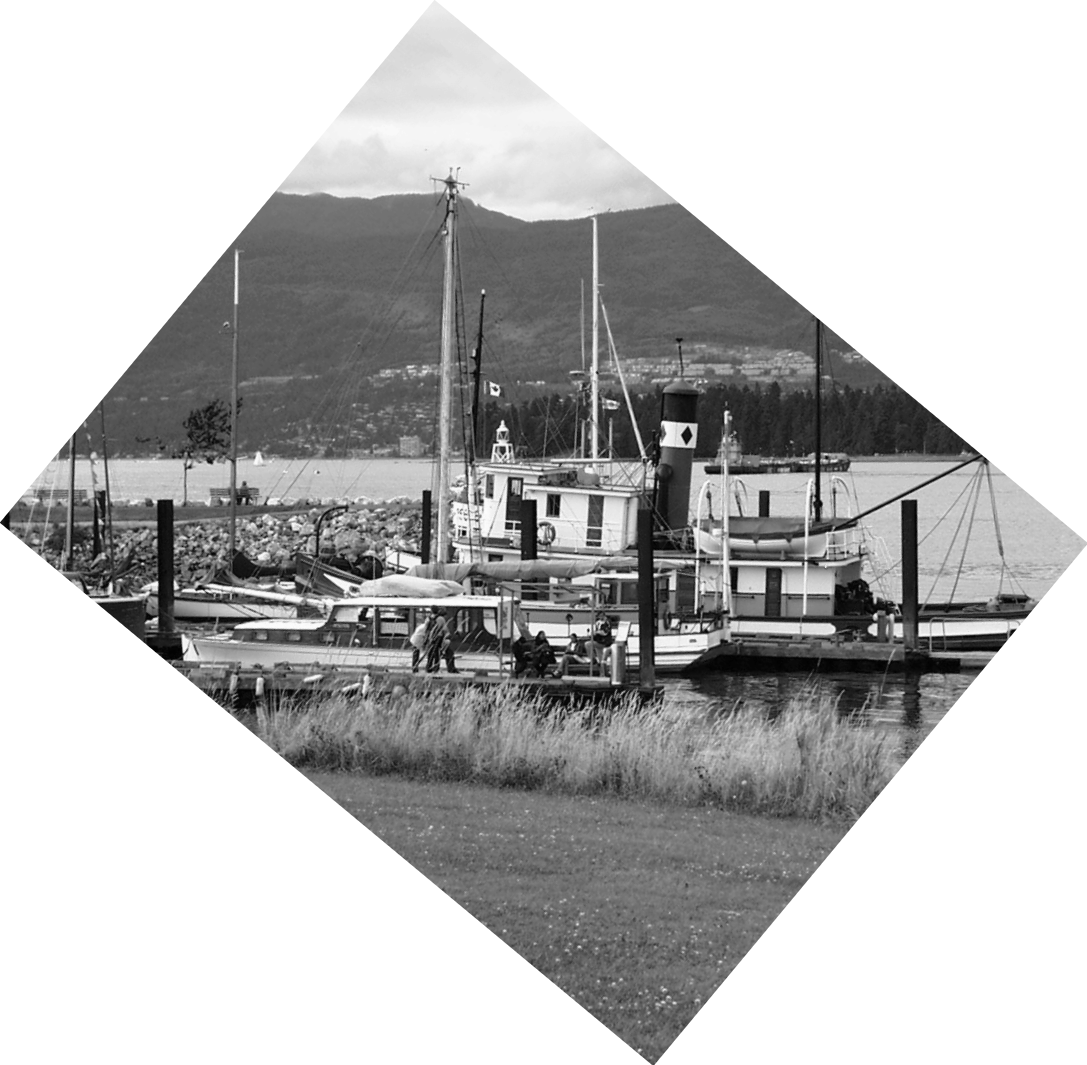}}} 
		\\[-\rowheight]
		\printcellmiddle
		& \printcellmiddle 
		& \printcellmiddle
		& \printcellmiddle \\
		\savecellbox{\rule{0pt}{4.073cm}}
		& \savecellbox{\subfigure[]{\includegraphics[height=3cm]{graf.png}}} \hspace{0.5cm}
		& \savecellbox{\subfigure[]{\includegraphics[height=3cm]{graf3.png}}} \hspace{0.5cm}
		& \savecellbox{\subfigure[]{\includegraphics[height=4.073cm]{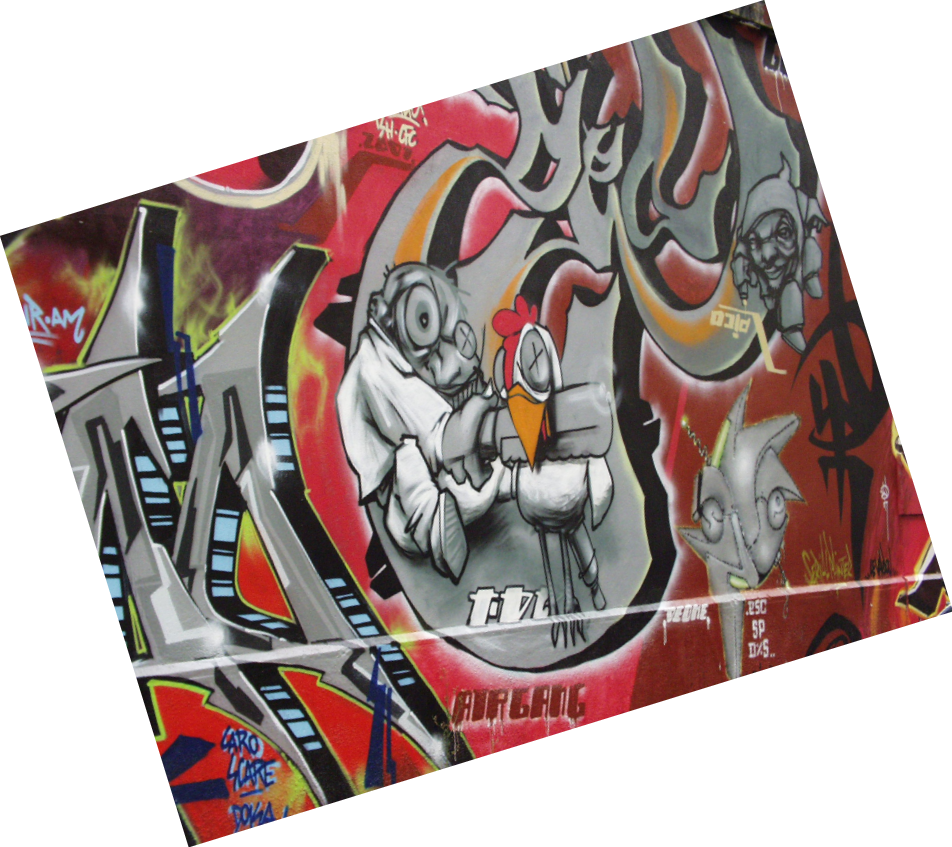}}}
		\\[-\rowheight]
		\printcellmiddle
		& \printcellmiddle 
		& \printcellmiddle
		& \printcellmiddle \\
	\end{tabular}
	\caption{Examples of image alignments: (a) \& (d) the first image, (b) \& (e) the second image, and (c) \& (f) the aligned second image.}
	\label{figure_images_with_rectification}
\end{figure*}
\indent Table \ref{table_true_matches} shows the number of true correspondences obtained by each approach. Notably, if we rely only on the similarity of feature descriptors, then the brute force approach should be able to obtain the optimal solution. However, the most similar descriptors do not necessarily indicate that the match is correct. If we can effectively limit the region of feature matching, thus avoiding the interference of matches from other impossible regions, we can sometimes achieve a better matching result. This can be observed from Table \ref{table_true_matches}, where our proposed system finds more correct matches than the brute force approach in several cases. Notably, the two cases in Figs. \ref{figure_Oxford_images}(c) and \ref{figure_Oxford_images}(f), both contain image rotations between the images to be matched. Since, in this case, the spatial order of features will change, which violates the assumption of the spatial order model, the numbers of true correspondences obtained by the SO and SO+EPI are much lower than those of the BF approach. However, after image alignment, our SO+EPI+IA can find more true correspondences than those of BF, indicating that our image alignment can effectively overcome the problem of image rotation. \\
\indent Figure \ref{figure_images_with_rectification} shows the results of image alignment of these two cases, where the first and second columns are the two images to be matched, and the third column shows the aligned images, which are transformed from the second column of Fig. \ref{figure_images_with_rectification}. Compared with the images in the first column of Fig. \ref{figure_images_with_rectification}, the images in the third column have been properly aligned, again confirming that our image alignment can effectively rectify the images. It is worth noting that the third column of Fig. \ref{figure_images_with_rectification} is for visually inspecting the effect of our image alignment. In practical applications, we do not really need to perform this alignment on the whole image, but can achieve good results by employing the transformation only on the coordinates of detected features. The computation of the spatial order model is based on the transformed coordinates, and the subsequent feature filtering is also performed according to the transformed coordinates. \\
\begin{table*}
	\footnotesize
	\centering
	\caption{Precisions for the images in Fig. \ref{figure_Oxford_images}.}
	\label{table:precision}
	\begin{tabular}{p{1.2cm}p{1.7cm}cccc} 
		\hline
		Image & Image Pair & BF & SO & SO+EPI & SO+EPI+IA \\ \hline
		\multirow{2}{1.2cm}{Fig. \ref{figure_Oxford_images}(a)} & img1 vs. img2 & 75.19$\%$ & 86.51$\%$ & 92.80$\%$ & 92.87$\%$\\ 
		& img1 vs. img3 & 70.76$\%$ & 82.96$\%$ & 91.25$\%$ & 91.33$\%$ \\ \hline
		\multirow{2}{1.2cm}{Fig. \ref{figure_Oxford_images}(b)} & img1 vs. img2 & 89.04$\%$ & 94.08$\%$ & 95.26$\%$ & 95.42$\%$\\ 
		& img1 vs. img3 & 84.82$\%$ & 91.79$\%$ & 96.26$\%$ & 96.32$\%$ \\ \hline
		\multirow{2}{1.2cm}{Fig. \ref{figure_Oxford_images}(c)} & img1 vs. img2 & 61.89$\%$ & 66.21$\%$ & 82.62$\%$ & 86.16$\%$\\ 
		& img1 vs. img3 & 51.81$\%$ & 55.97$\%$ & 80.00$\%$ & 84.23$\%$ \\ \hline
		\multirow{2}{1.2cm}{Fig. \ref{figure_Oxford_images}(d)} & img1 vs. img2 & 73.95$\%$ & 82.37$\%$ & 89.74$\%$ & 89.82$\%$\\ 
		& img1 vs. img3 & 65.51$\%$ & 76.82$\%$ & 84.74$\%$ & 84.74$\%$ \\ \hline
		\multirow{2}{1.2cm}{Fig. \ref{figure_Oxford_images}(e)} & img1 vs. img2 & 92.07$\%$ & 97.76$\%$ & 98.16$\%$ & 98.16$\%$\\ 
		& img1 vs. img3 & 88.04$\%$ & 95.66$\%$ & 96.78$\%$ & 96.67$\%$ \\ \hline
		\multirow{2}{1.2cm}{Fig. \ref{figure_Oxford_images}(f)} & img1 vs. img2 & 38.49$\%$ & 34.10$\%$ & 70.20$\%$ & 71.58$\%$\\ 
		& img1 vs. img3 & 14.32$\%$ & 16.62$\%$ & 49.65$\%$ & 46.06$\%$ \\ \hline
	\end{tabular}
	\label{table_precision}
\end{table*}
\indent Table \ref{table_precision} shows the precisions obtained by the brute force approach and by our framework. Precision is calculated by dividing the number of true correspondences by the number of matched features. As can be seen from the table, SO significantly outperforms BF, indicating that our SO feature filtering is effective in removing incorrect matches. The precision of SO+EPI is even higher, showing a better effect of filtering out incorrect matches. The precision obtained by the SO+EPI+IA is comparable to that of SO+EPI, meaning that the inclusion of image alignment does not degrade the precision. From Tables \ref{table_true_matches} and \ref{table_precision}, we observe that, compared with the brute force approach, our guided image feature matching not only does not reduce the detected true correspondences, but also, in many cases, increases the detected true correspondences with improved precision. \\
\begin{table*}
	\footnotesize
	\centering
	\caption{Matching time (seconds) for the images in Fig. \ref{figure_Oxford_images}.}
	\begin{tabular}{p{1.2cm}p{1.7cm}cccc} 
		\hline
		Image & Image Pair & BF & SO & SO+EPI & SO+EPI+IA \\ \hline
		\multirow{2}{1.2cm}{Fig. \ref{figure_Oxford_images}(a)} & img1 vs. img2 & 0.41902 & 0.11997 & 0.05825 & 0.05609 \\ 
		& img1 vs. img3 & 0.37375 & 0.11906 & 0.05506 & 0.05556 \\ \hline
		\multirow{2}{1.2cm}{Fig. \ref{figure_Oxford_images}(b)} & img1 vs. img2 & 1.66756 & 0.45064 & 0.13545 & 0.13658 \\ 
		& img1 vs. img3 & 1.64142 & 0.72711 & 0.16079 & 0.16164 \\ \hline
		\multirow{2}{1.2cm}{Fig. \ref{figure_Oxford_images}(c)} & img1 vs. img2 & 0.91389 & 0.45799 & 0.13951 & 0.09918 \\ 
		& img1 vs. img3 & 0.75956 & 0.38946 & 0.10548 & 0.10435 \\ \hline
		\multirow{2}{1.2cm}{Fig. \ref{figure_Oxford_images}(d)} & img1 vs. img2 & 0.37882 & 0.11318 & 0.04485 & 0.04583 \\ 
		& img1 vs. img3 & 0.33706 & 0.10571 & 0.04236 & 0.04539 \\ \hline
		\multirow{2}{1.2cm}{Fig. \ref{figure_Oxford_images}(e)} & img1 vs. img2 & 0.63540 & 0.12660 & 0.06193 & 0.06390 \\ 
		& img1 vs. img3 & 0.64059 & 0.12875 & 0.06435 & 0.06669 \\ \hline
		\multirow{2}{1.2cm}{Fig. \ref{figure_Oxford_images}(f)} & img1 vs. img2 & 0.49798 & 0.31581 & 0.12527 & 0.08780 \\ 
		& img1 vs. img3 & 0.55405 & 0.38560 & 0.14203 & 0.13976 \\ \hline
	\end{tabular}
	\label{table_matching_time}
\end{table*}
\indent Table \ref{table_matches} confirms that our framework is indeed effective in reducing the number of matches. This reduction means that our framework is more efficient in feature matching. Table \ref{table_matching_time} reveals the required matching times of our framework and the brute force approach on the test image sets in Fig. \ref{figure_Oxford_images}. Note that this time does not include the time of feature detection. Since the computer takes a slightly different time for each execution, we conducted twenty iterations of each method for each experimental set, and Table \ref{table_matching_time} lists the average times for these twenty trials. Table 4 shows that the time required for SO was approximately 20\% to 70\% of that of the BF approach, the time required for SO+EPI is approximately 8\% to 26\% of the BF, and the time required for SO+EPI+IA was comparable to that of SO+EPI. Notably, since the estimations of the fundamental matrix and spatial order model take some time, the time required for feature matching is not exactly in proportion to the number of matches. \\
\indent The experimental results of the Oxford image dataset demonstrate that our framework can effectively reduce the number of matches, effectively improve the precision of feature matching, in many cases increase the number of detected true correspondences, and effectively improve matching efficiency. \\

\subsection{Simulated Images}
\indent To further validate the effectiveness of our framework, we generated some simulated images for testing. We selected 100 images from the ImageNet dataset, and performed image transformations on them to generate the images to be matched. The image transformations we applied were based on \cite{Dosovitskiy16} with slight modifications, and are summarized as follows: \\
\begin{itemize}
	\item Rotation: Rotate the image around its center with an angle between $-90^\circ$ and $90^\circ$.
	\item Translation: Translate the image by a distance within 0.1 of the image size.
	\item Scaling: Scale the image by a factor between 0.5 and 2.0.
	\item Contrast variation 1: Multiply the projection of each pixel onto the principal component of the set of all pixels by a factor between 0.8 and 1.2.
	\item Contrast variation 2: Transform the image to the hue, saturation, value (HSV) color representation and then raise saturation and value of all pixels to a power between 0.5 and 2. Multiply these values by a factor between 0.8 and 1.2, and add to them a value between $-0.05$ and $0.05$.
	\item Hue variation: Add a value between $-0.05$ and $0.05$ to the hue of all pixels in the image.
	\item Perspective transformation: Randomly add a value of 0.05 of the image size to the four corners of the image. Then, according to the modified four corners, compute a homography matrix to transform the image.
\end{itemize}
Note that, compared with \cite{Dosovitskiy16}, we have included a new transformation, \textit{i.e.}, the perspective transformation, to enrich image variations of our test set. \\
\begin{figure*}
	\centering
	\begin{tabular}[t]{cccc}
		\subfigure[2400$\times$1600]{\includegraphics[width=2.67cm]{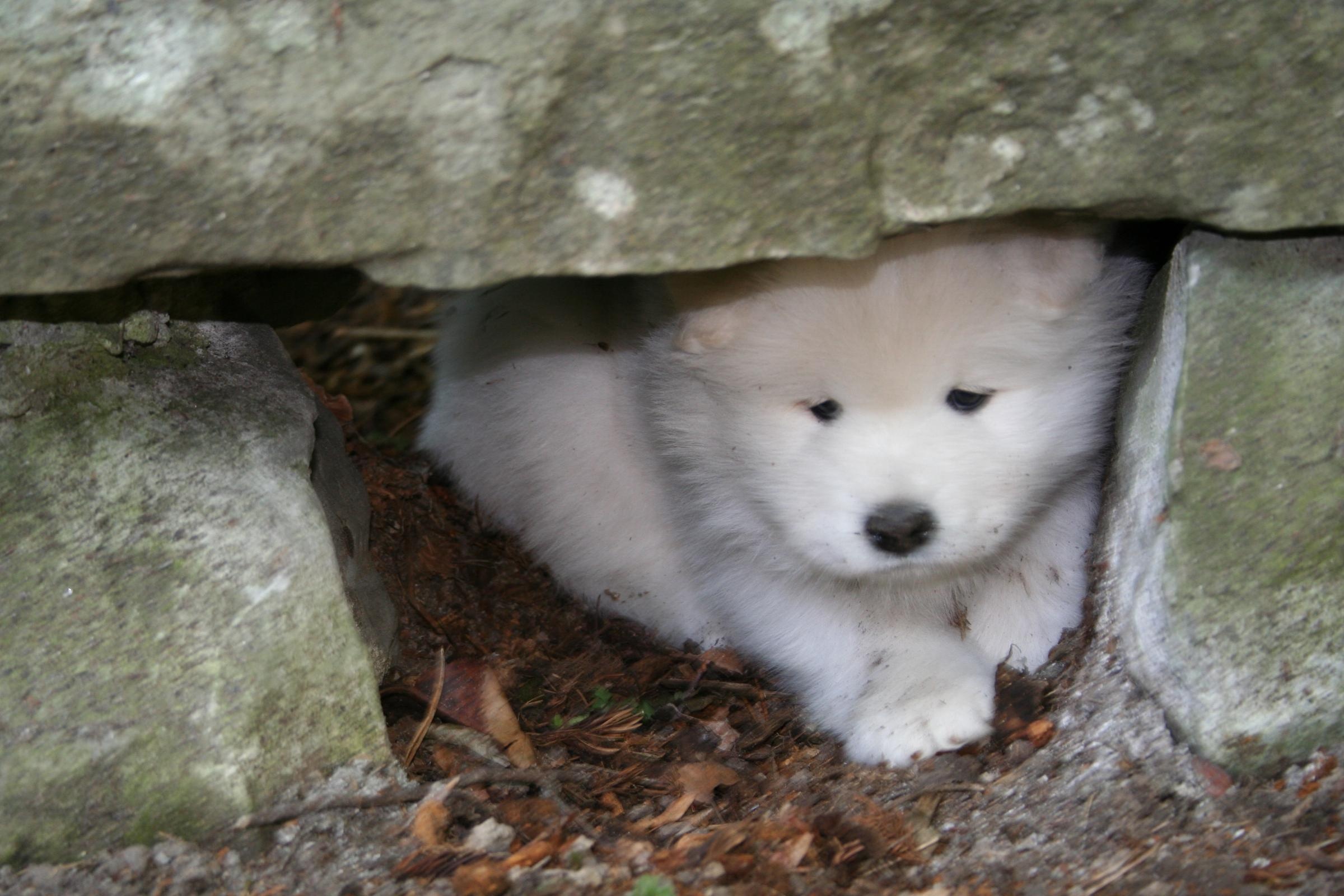}} \hspace{0.1cm}
		\subfigure[2048$\times$1536]{\includegraphics[width=2.27cm]{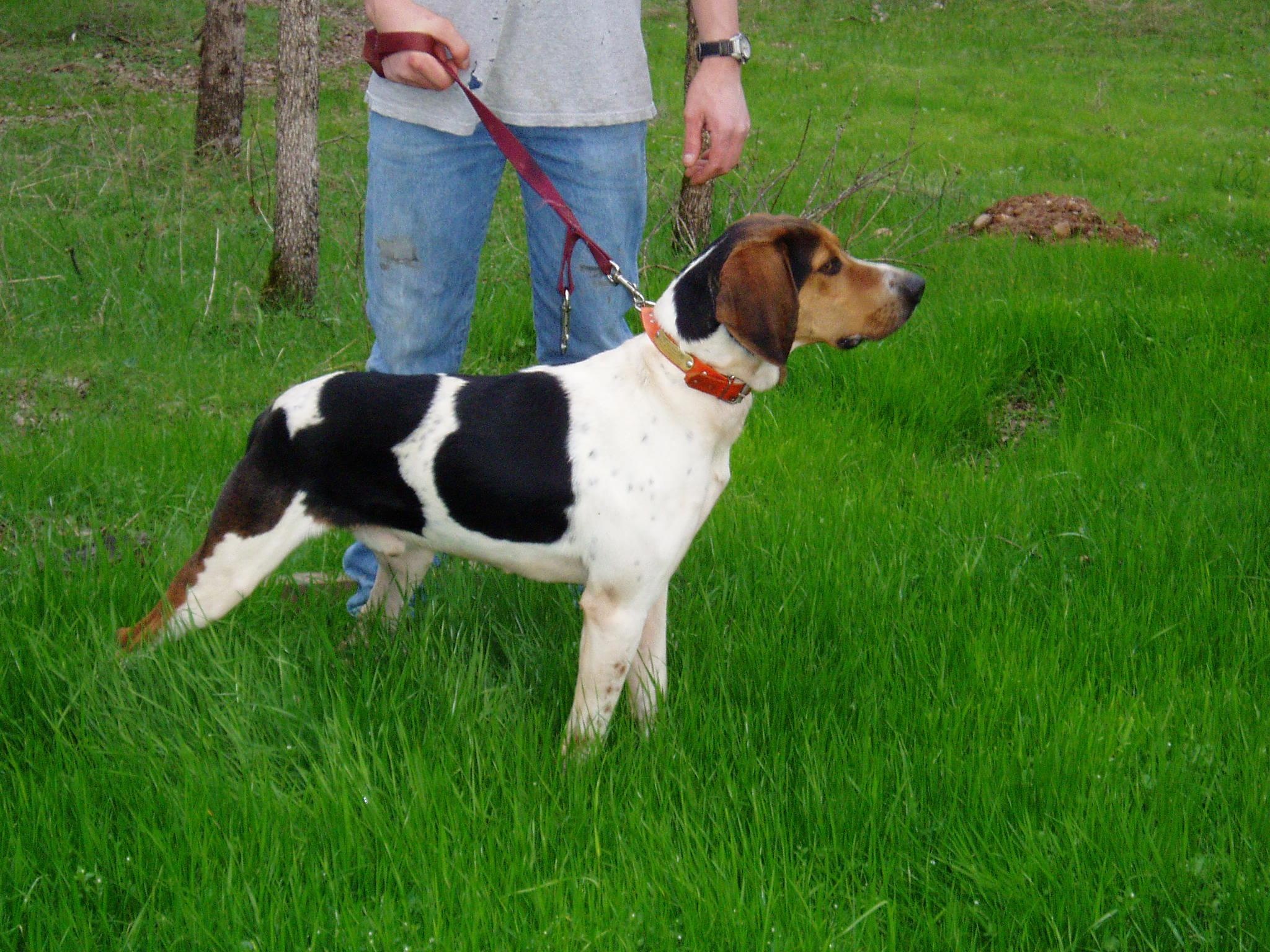}} \hspace{0.1cm}
		\subfigure[2136$\times$2848]{\includegraphics[width=2.37cm]{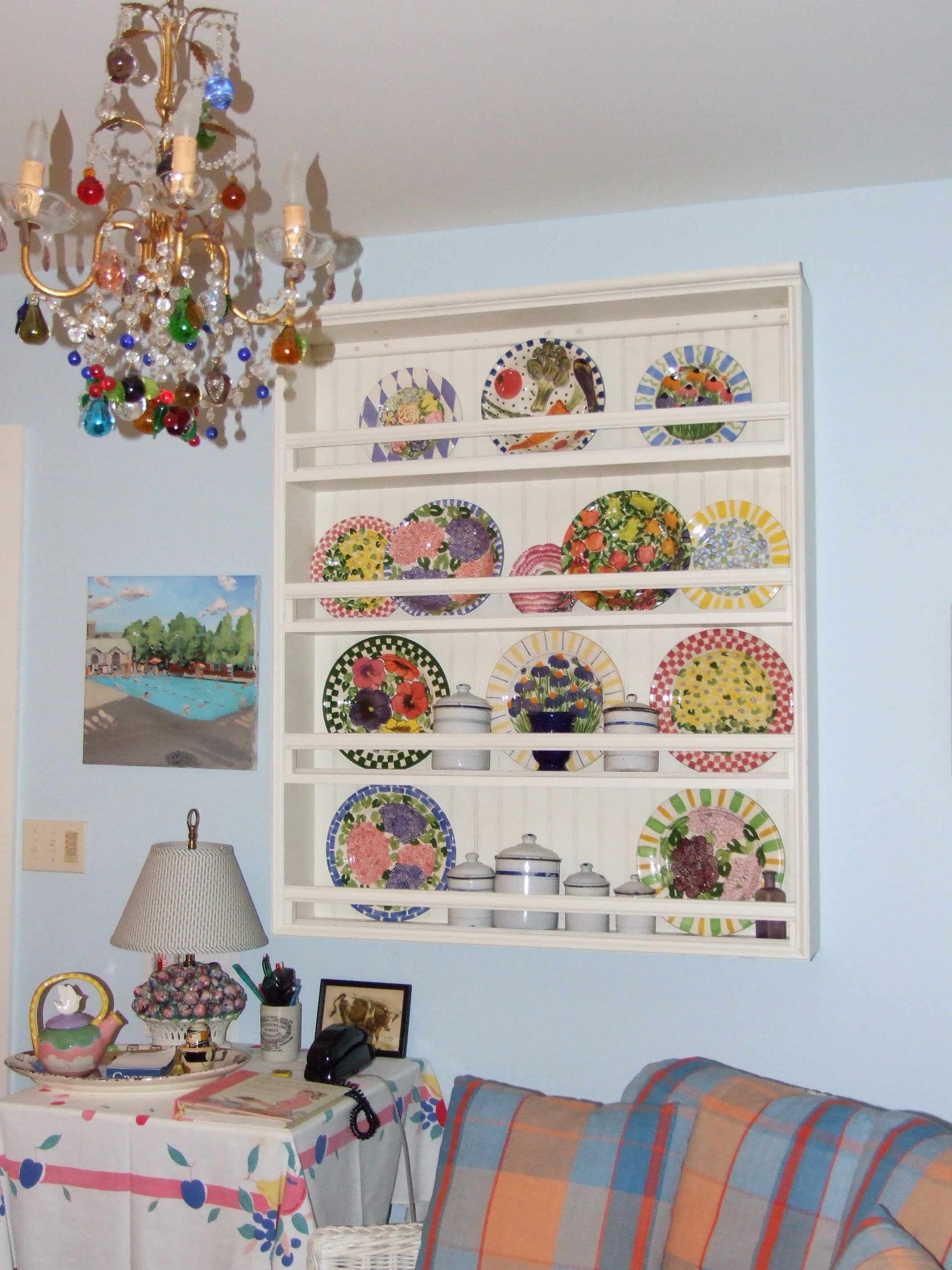}} \\
		\subfigure[1542$\times$1721]{\includegraphics[width=1.71cm]{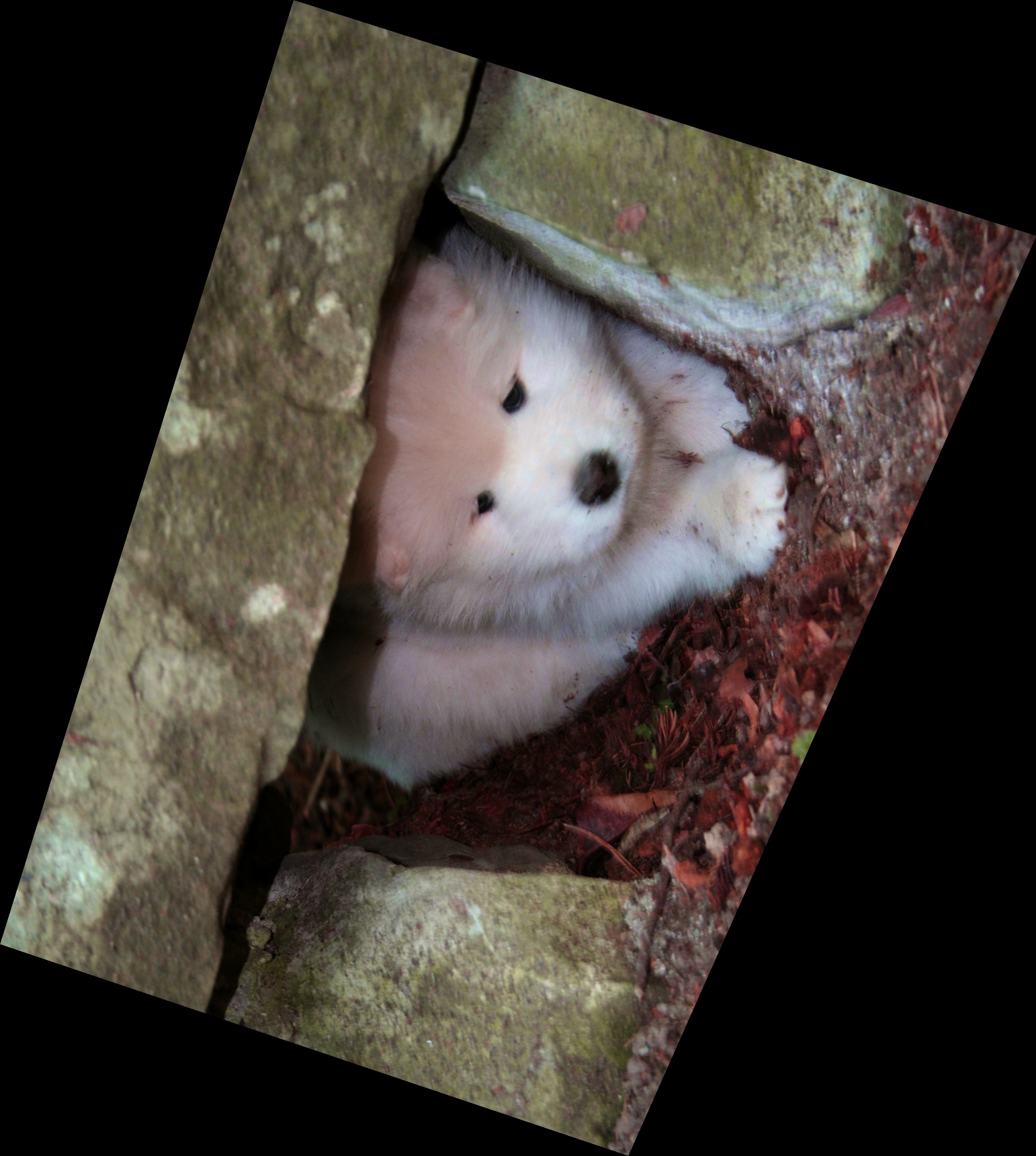}} \hspace{0.1cm}
		\subfigure[2606$\times$2457]{\includegraphics[width=2.89cm]{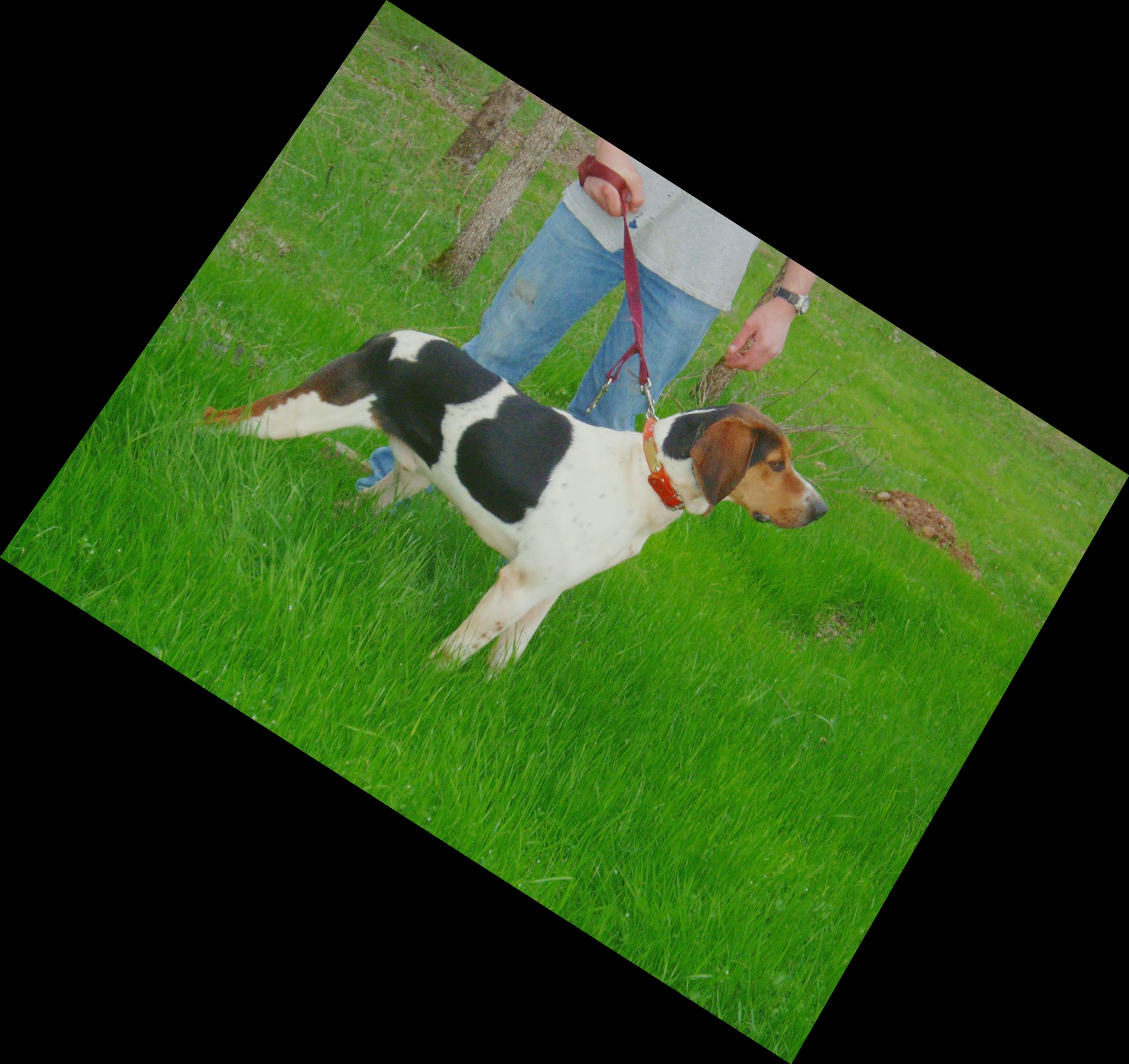}} \hspace{0.1cm}
		\subfigure[3367$\times$3023]{\includegraphics[width=3.74cm]{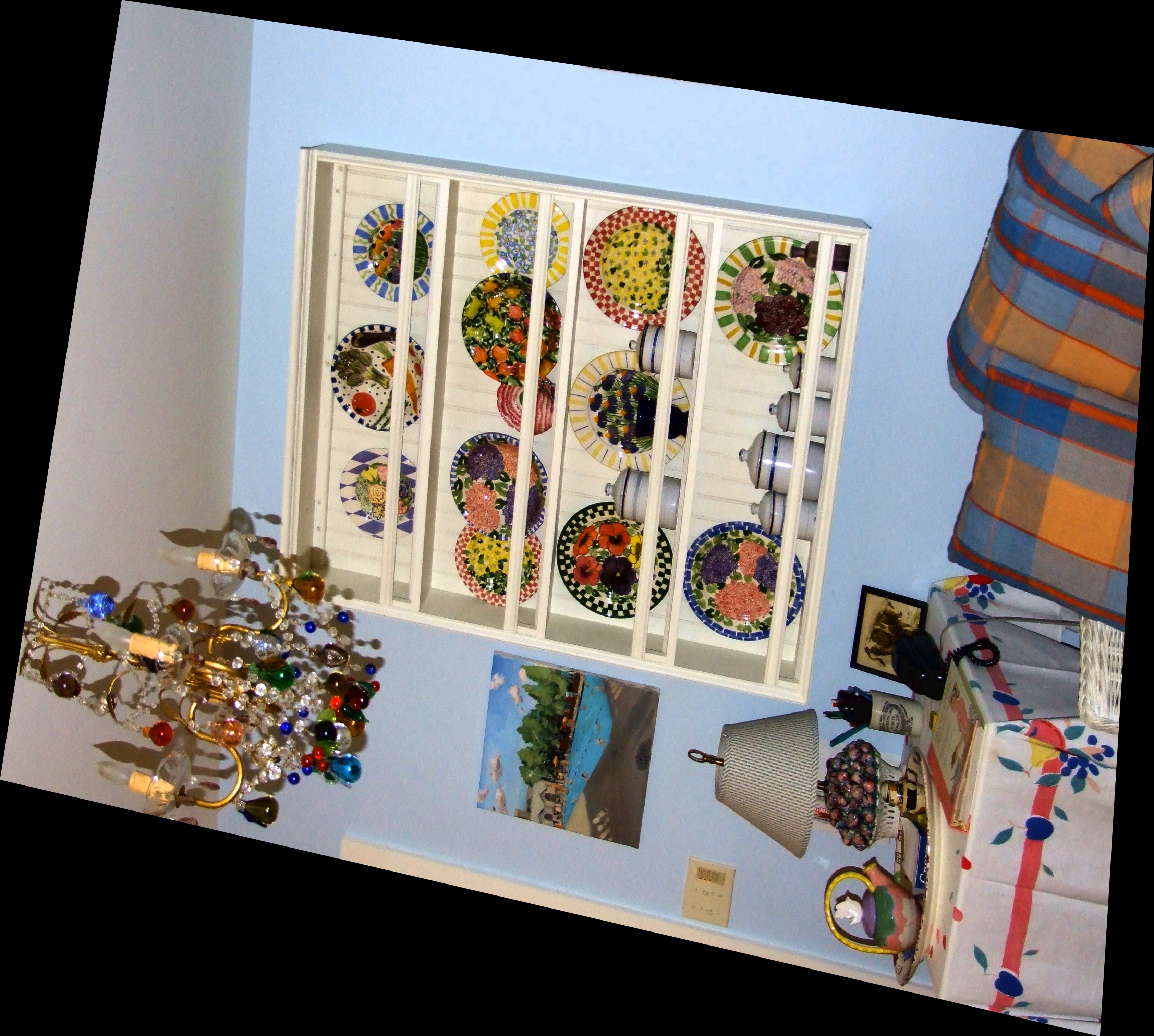}} \\
		\subfigure[2091$\times$1971]{\includegraphics[width=2.32cm]{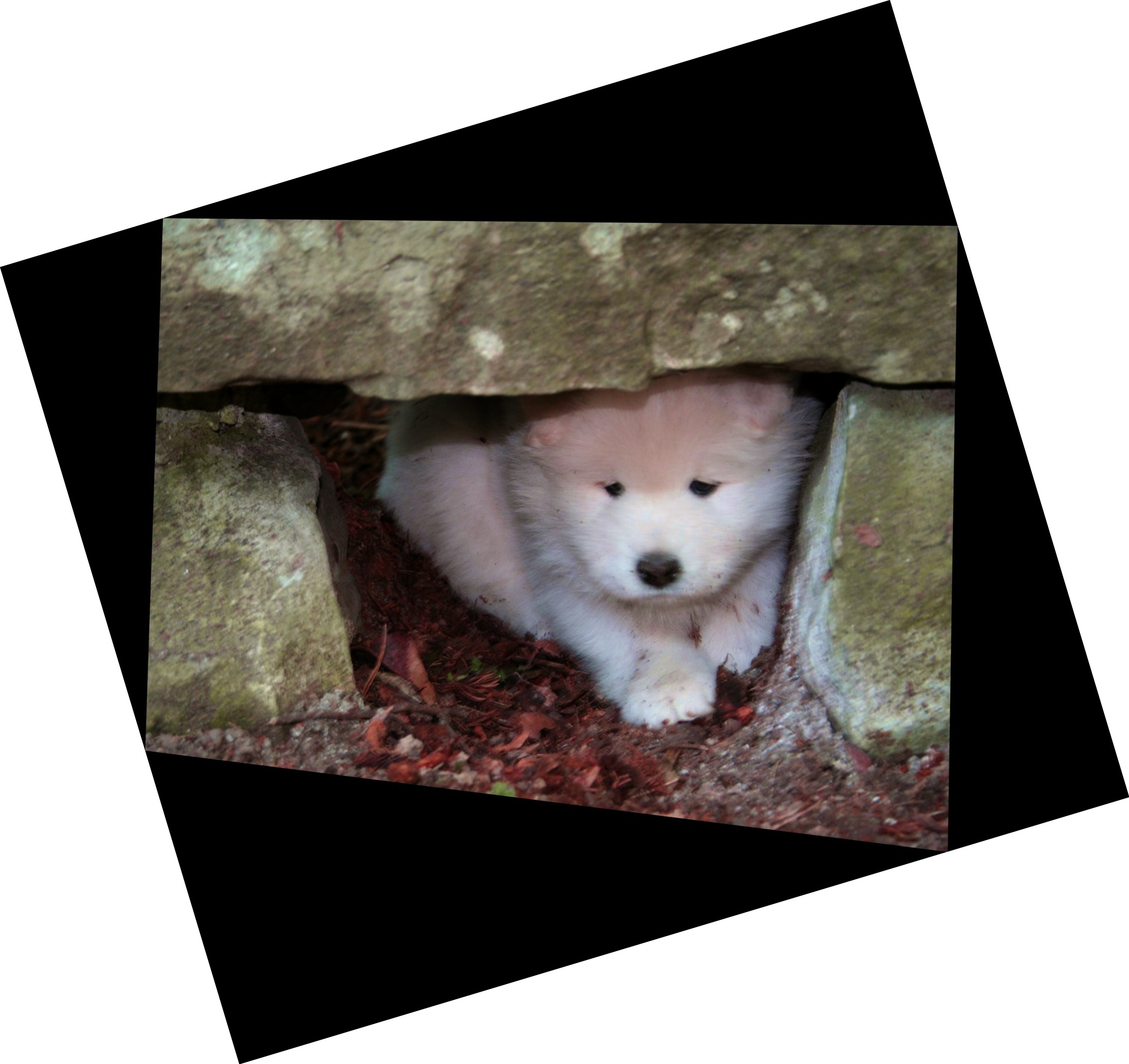}} \hspace{0.1cm}
		\subfigure[3516$\times$3470]{\includegraphics[width=3.90cm]{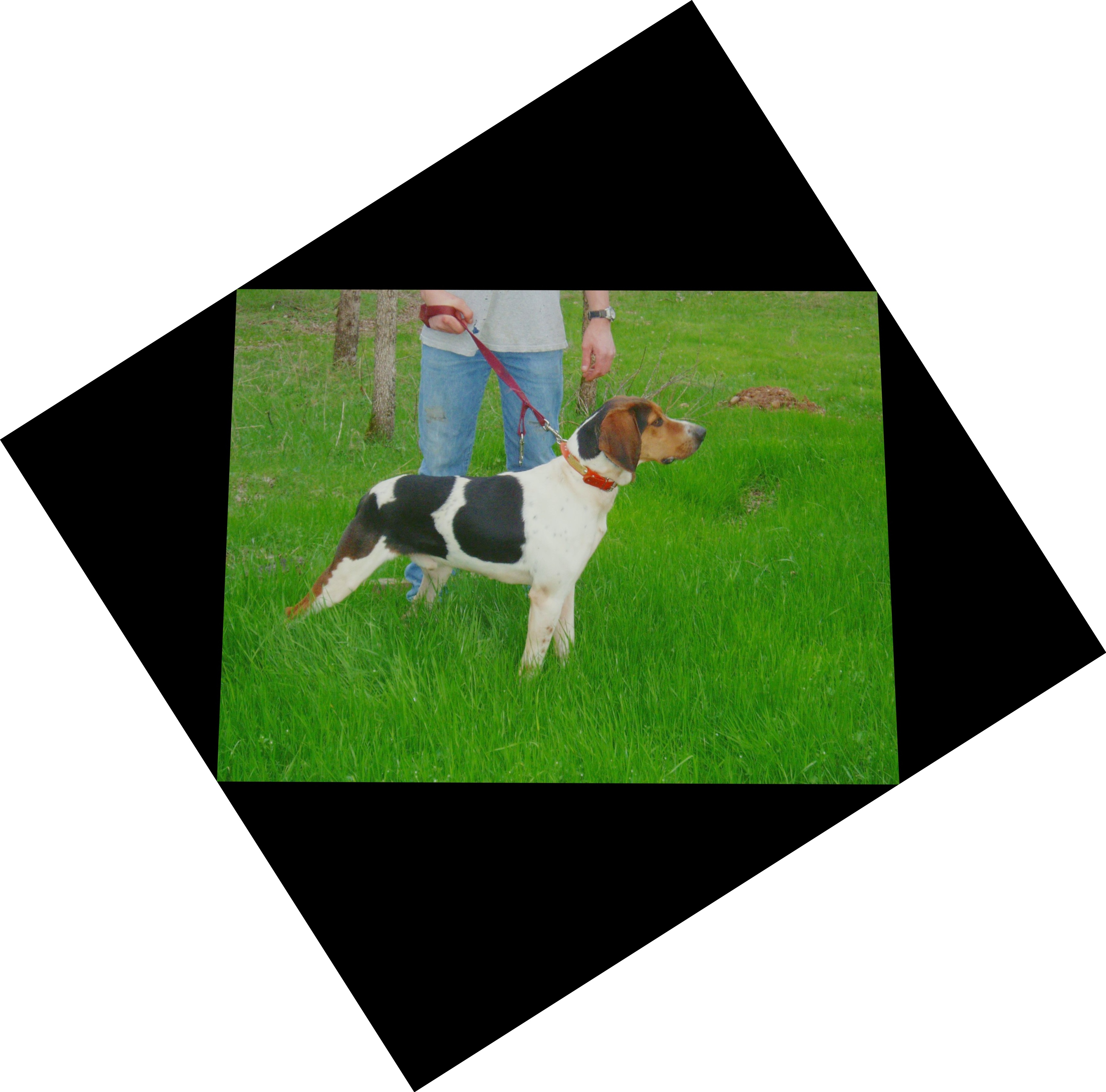}} \hspace{0.1cm}
		\subfigure[3492$\times$3781]{\includegraphics[width=3.87cm]{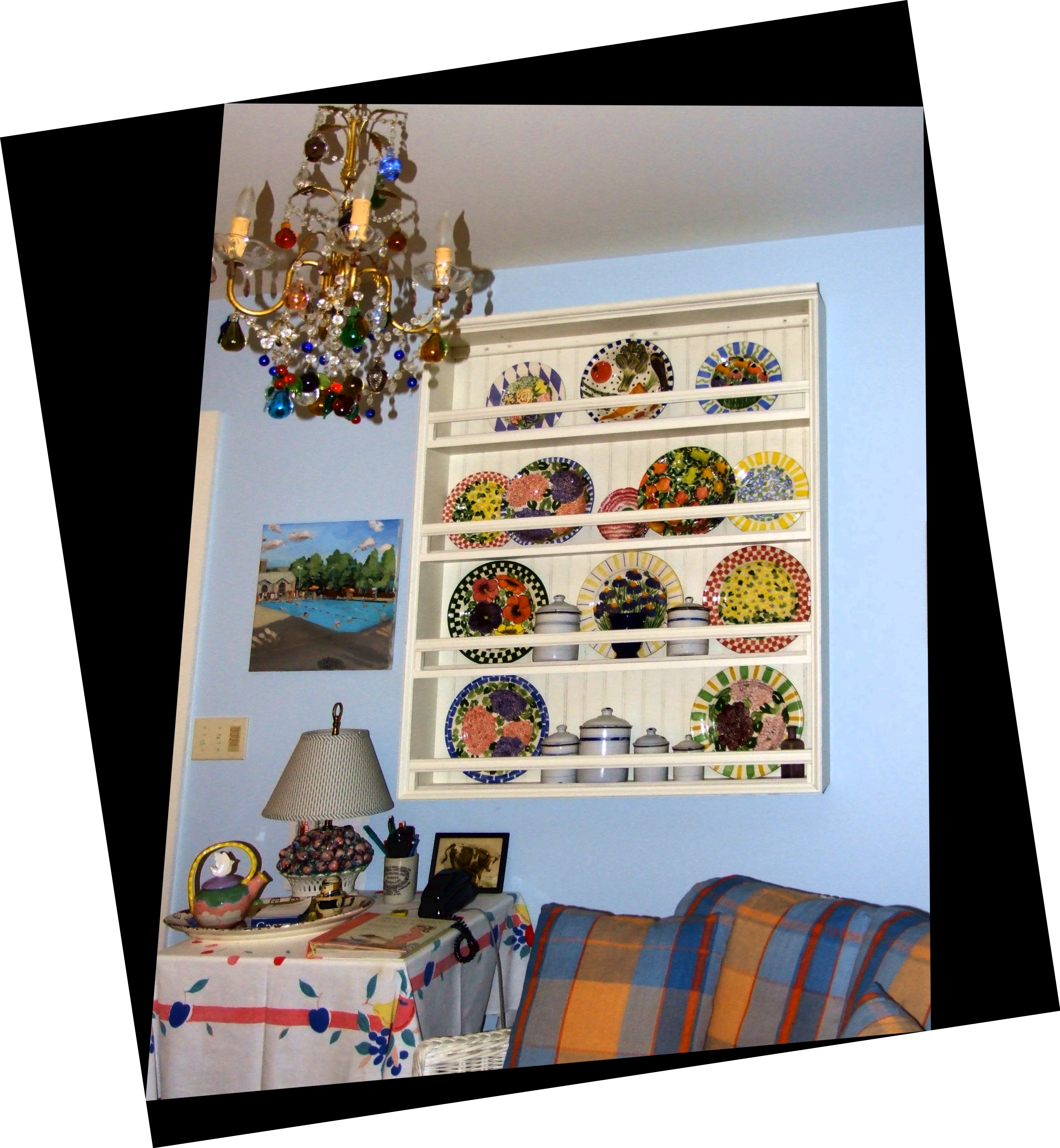}} \\
	\end{tabular}
	\caption{Some images from ImageNet. The upper row is the original image, the middle row is the transformed image (\textit{i.e.}, the image to be matched), and the bottom row is the aligned image.}
	\label{figure_simulated_images}
\end{figure*}
\begin{table*}
	\footnotesize
	\centering
	\caption{Experimental results for 100 test images with color transformation, translation, and scaling.}
	\begin{tabular}{p{2cm}cccc} 
		\hline
		& \# matches     & \# true correspondences & precision & matching time (s) \\ \hline
		BF         & 71,669,608,993 & 597,597                 & 61.66\%   & 2368.7 \\ 
		SO         & 8,584,122,656  & 613,811                 & 72.38\%   & 511.5 \\ 
		SO+EPI     & 336,334,792    & 633,598                 & 86.21\%   & 93.9 \\ 
		SO+EPI+IA  & 336,529,998    & 633,199                 & 86.19\%   & 95.5 \\ \hline
	\end{tabular}
	\label{table_simulated_images_set1}
\end{table*}
\indent The first row of Fig. \ref{figure_simulated_images} shows three of the 100 images we selected, and the second row displays the corresponding transformed images to be matched. To verify the effect of our method under different image geometry transformations, we generated three datasets from these 100 images. In additional to the color transformations, the first dataset contains only translation and scaling in geometry transformations. Table \ref{table_simulated_images_set1} lists the experimental results of this dataset. The data listed in the table are the sums of the values of these 100 images, except for precision, which is the overall precision of these 100 images. \\
\indent As can be seen from the table, SO can reduce the number of matches to approximately 12\% of that of BF, while SO+EPI can further reduce the number of matches to approximately 0.5\% of that of BF. The number of matches for SO+EPI+IA is comparable to that of SO+EPI. Since the test set is generated according to known geometric transformations, we can know whether a match is correct or not. As can be seen in Table \ref{table_simulated_images_set1}, SO, SO+EPI, and SO+EPI+IA all yield more true correspondences than BF. In terms of matching accuracy, all three of our approaches can obtain a higher precision than the BF method. Finally, the matching time required for SO is approximately 22\% of the BF, the matching time required for SO+EPI is approximately 4\% of the BF, and the matching time required for SO+EPI+IA is slightly higher than that for SO+EPI. It is worth noting that, although the computation of the fundamental matrix and the spatial order model requires some time, our framework can obtain significant performance improvement for images with a large number of features detected. \\
\begin{table*}
	\footnotesize
	\centering
	\caption{Experimental results for 100 test images with color transformation, scaling, and rotation.}
	\begin{tabular}{p{2cm}cccc} 
		\hline
		& \# matches     & \# true correspondences & precision & matching time (s) \\ \hline
		BF         & 61,477,355,511 & 351,920                 & 44.74\%   & 2069.5 \\ 
		SO         & 26,354,403,902 & 181,311                 & 37.57\%   & 1009.4 \\ 
		SO+EPI     & 428,625,894    & 191,547                 & 73.40\%   & 99.5 \\ 
		SO+EPI+IA  & 361,549,238    & 379,939                 & 80.21\%   & 86.8 \\ \hline
	\end{tabular}
	\label{table_simulated_images_set2}
\end{table*}
\indent Table \ref{table_simulated_images_set2} shows the experimental results of the second dataset, in which images contained scaling and rotation geometric transformations. Because image rotation can cause a change in the spatial order of features, the spatial order model may not perform well with image rotation. This is verified by Table \ref{table_simulated_images_set2} where the number of true correspondences obtained by the SO and SO+EPI is much smaller than that of BF, and the precision of SO is also lower than that of BF. However, after image alignment (\textit{i.e.}, SO+EPI+IA), we find that the produced number of true correspondences is higher than that of BF and its precision is also higher than that of the other three approaches. This is because the SO model can be built correctly after image alignment. The number of matches and the matching time of SO+EPI+IA are the lowest among the four methods. These results demonstrate that our image alignment can be integrated with SO+EPI to overcome the problem of image rotation. \\
\begin{table*}
	\footnotesize
	\centering
	\caption{Experimental results for 100 test images with color transformation, scaling, rotation, and perspective transformation.}
	\begin{tabular}{p{2cm}cccc} 
		\hline
		& \# matches     & \# true correspondences & precision & matching time (s) \\ \hline
		BF         & 71,386,852,178 & 351,986                 & 45.75\%   & 2466.5 \\ 
		SO         & 27,152,940,520 & 182,657                 & 38.39\%   & 1082.9 \\ 
		SO+EPI     & 529,620,879    & 190,423                 & 72.59\%   & 114.0 \\ 
		SO+EPI+IA  & 484,626,804    & 331,041                 & 79.09\%   & 103.1 \\ \hline
	\end{tabular}
	\label{table_simulated_images_set3}
\end{table*}
\indent Our final dataset adds a perspective transformation to the geometric transformations used in the previous dataset. From the experimental results shown in Table \ref{table_simulated_images_set3}, we can see that SO+EPI+IA can still effectively match image features. Its matching speed is greatly improved, and its precision is also better than that of BF, except where the number of detected true correspondences is slightly lower than that of BF. This indicates that, although our image alignment can overcome the problem of image rotation, its performance is slightly degraded under the condition of perspective distortion. However, if the matching accuracy, efficiency, and the number of detected true correspondences are comprehensively considered, then SO+EPI+IA is still a method with practical application value. The third row of Fig. \ref{figure_simulated_images} shows the results of the images in the second row after image alignment. Note that the rightmost image in Fig. \ref{figure_simulated_images} has a serious perspective distortion, but our image alignment manages to transform it to a proper orientation.
\subsection{Real Images}
\begin{figure*}
	\centering
	\begin{tabular}[t]{cccc}
		\addtocounter{subfigure}{-4}
		\subfigure{\includegraphics[height=3cm]{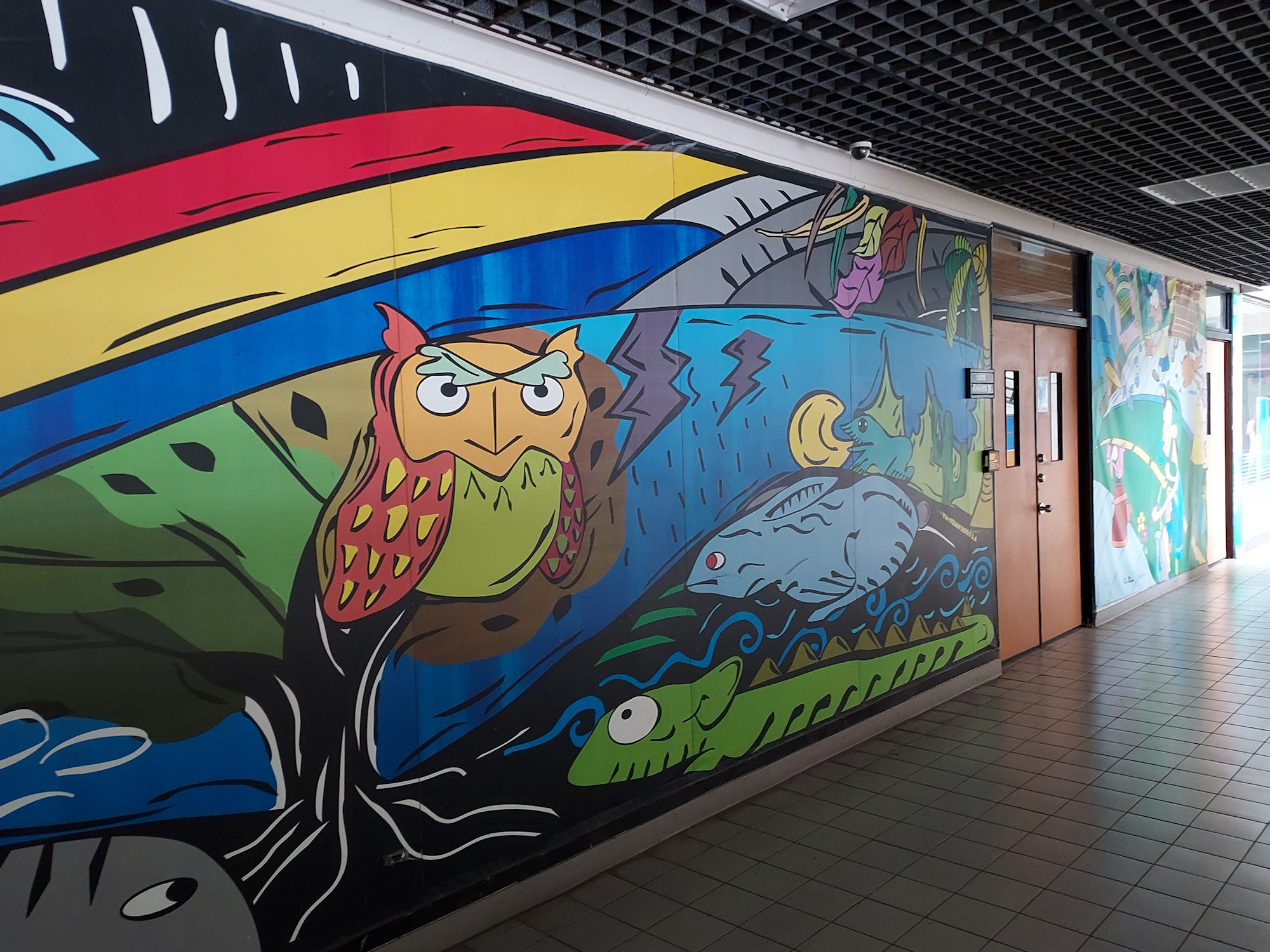}} \hspace{0.1cm}
		\subfigure{\includegraphics[height=3cm]{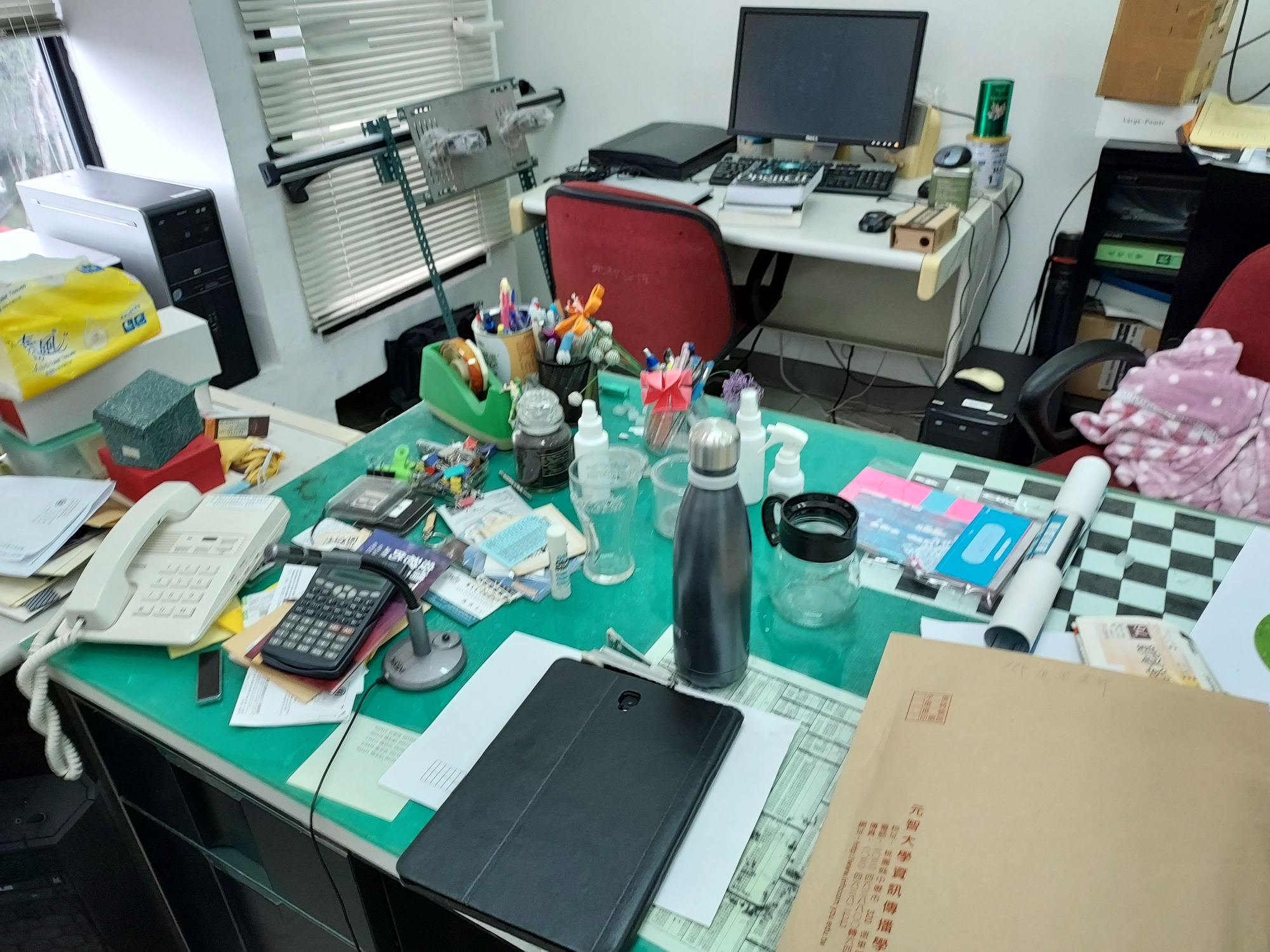}} \hspace{0.1cm}
		\subfigure{\includegraphics[height=3cm]{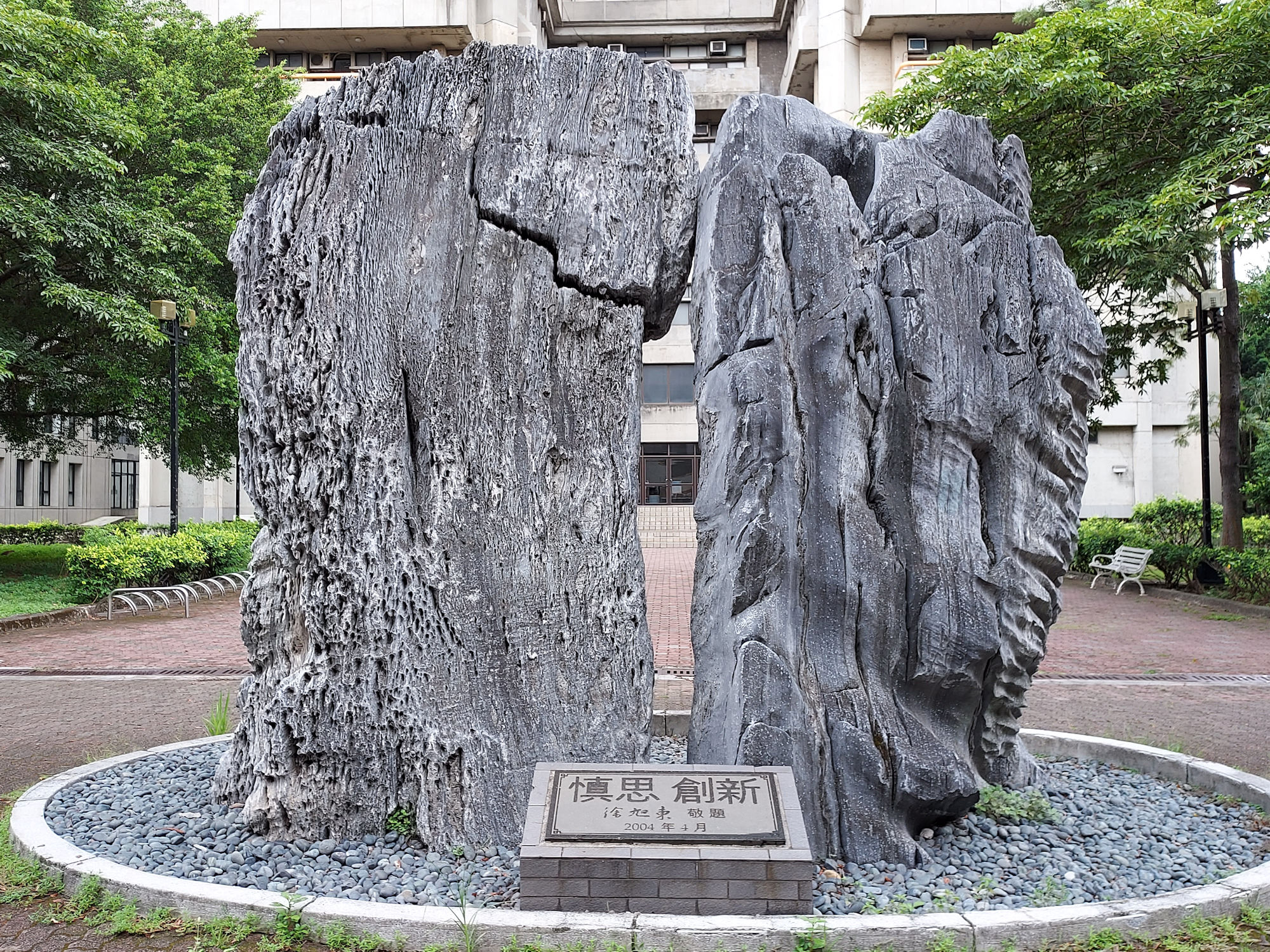}} \hspace{0.1cm}
		\subfigure{\includegraphics[height=3cm]{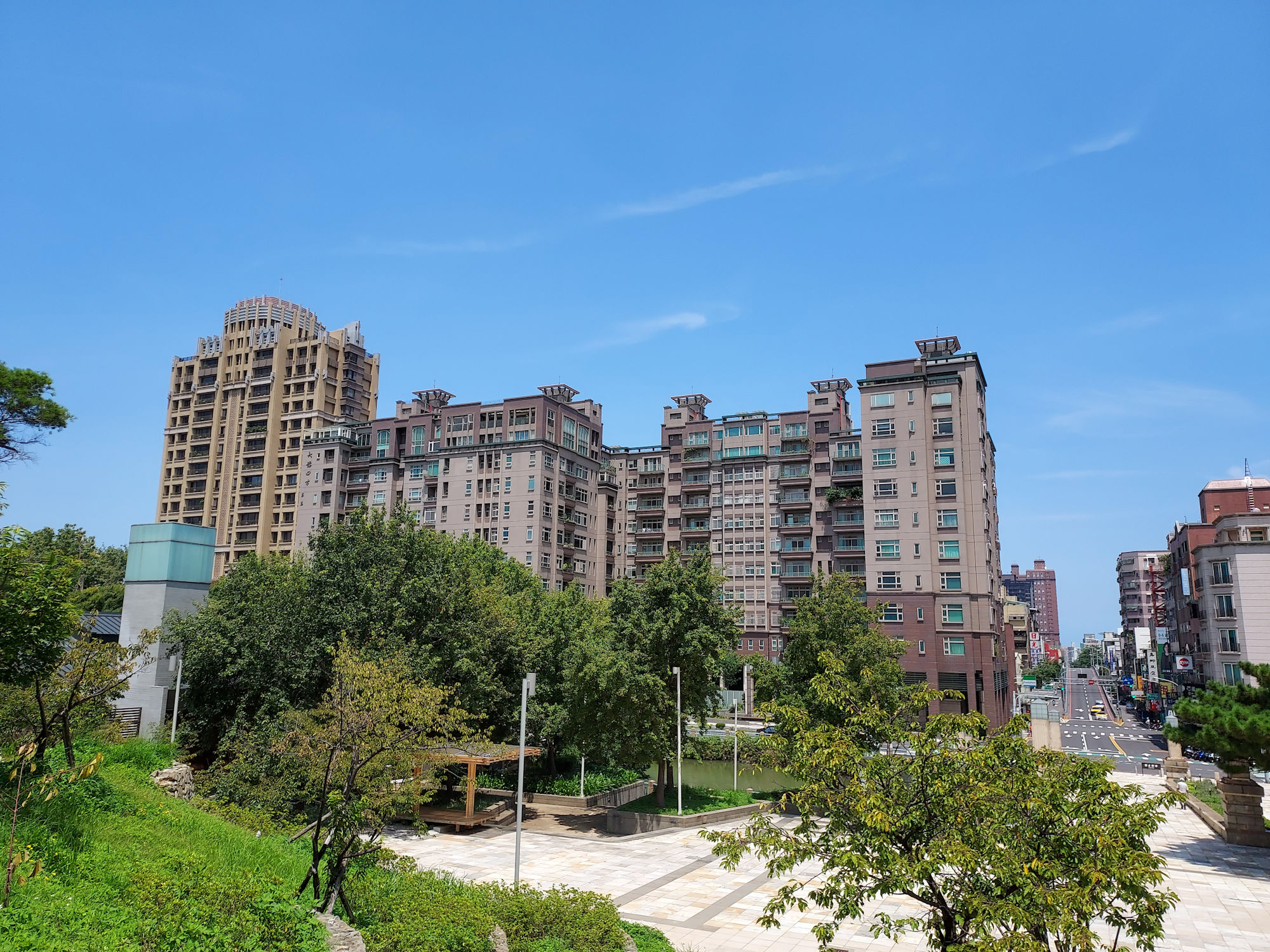}} \\
		\subfigure[]{\includegraphics[height=3cm]{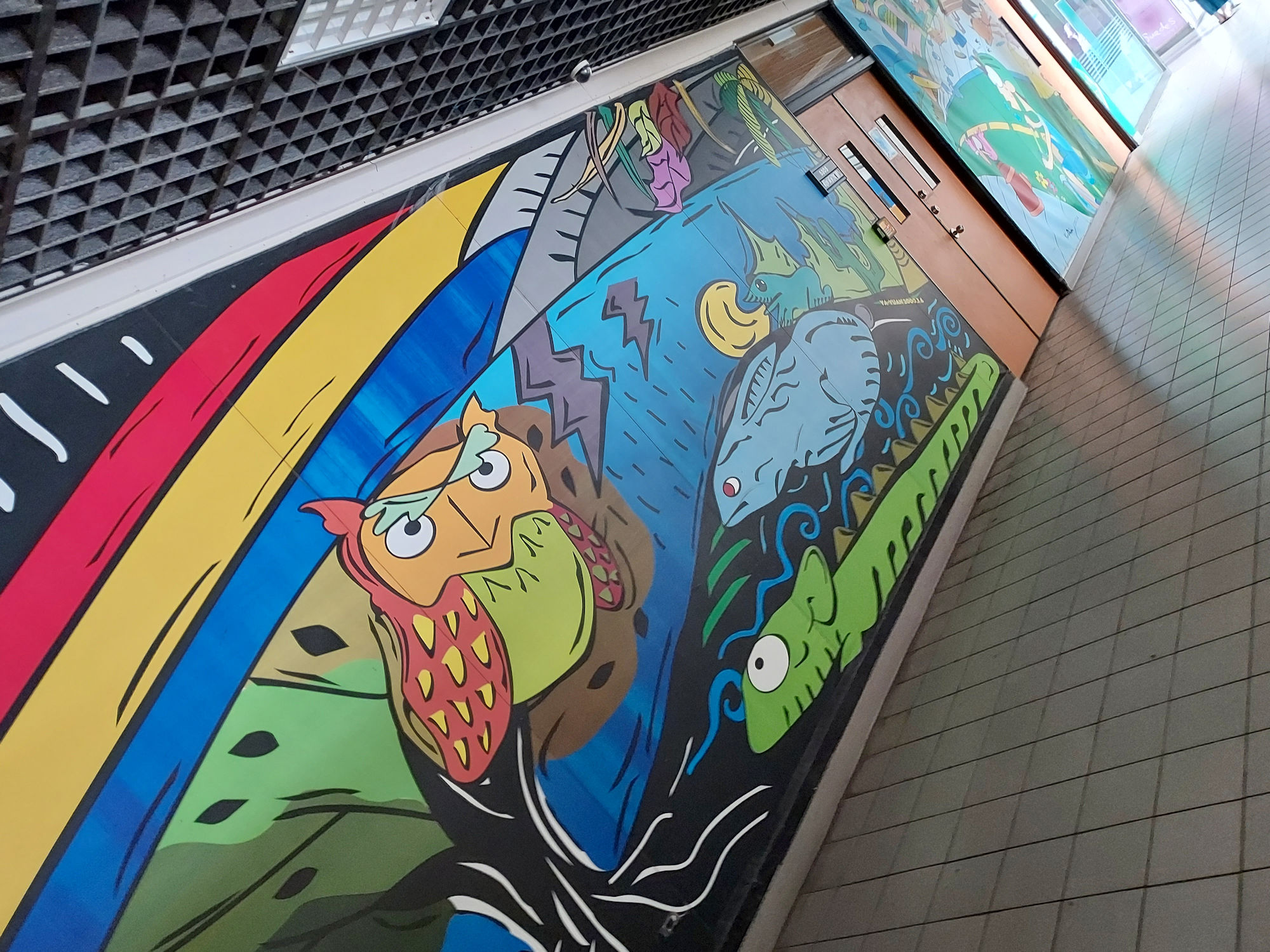}} \hspace{0.1cm}
		\subfigure[]{\includegraphics[height=3cm]{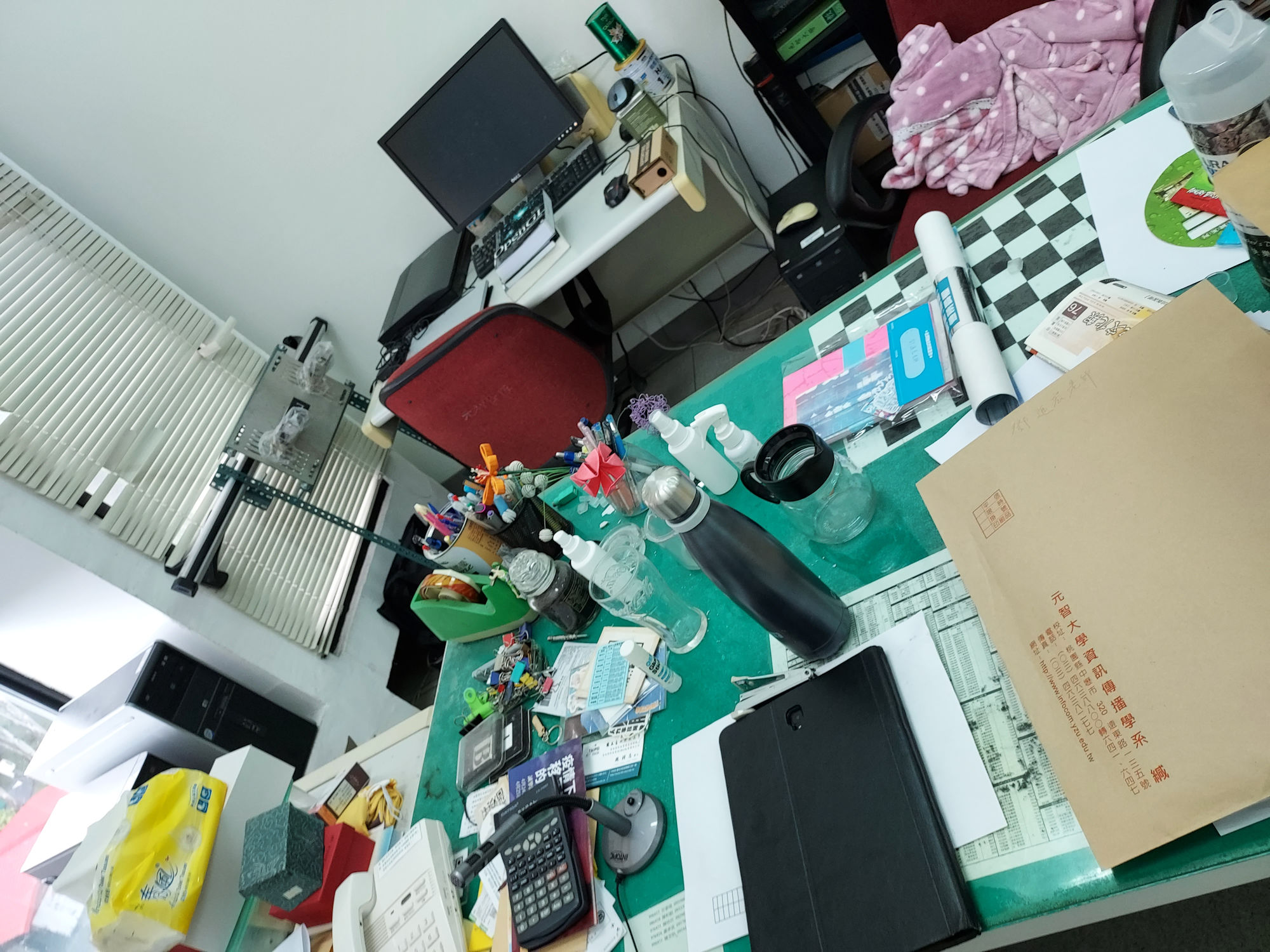}} \hspace{0.1cm}
		\subfigure[]{\includegraphics[height=3cm]{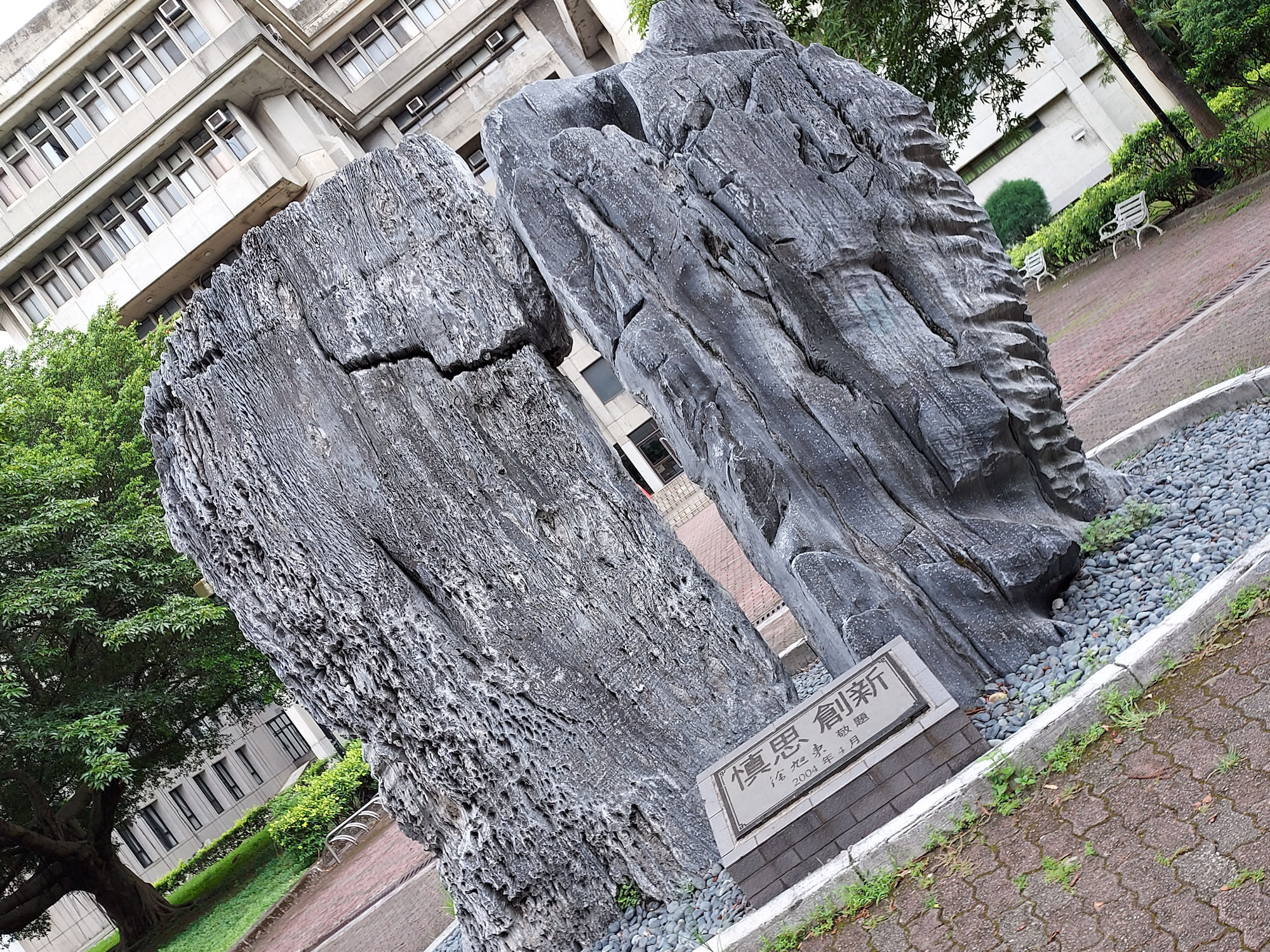}} \hspace{0.1cm}
		\subfigure[]{\includegraphics[height=3cm]{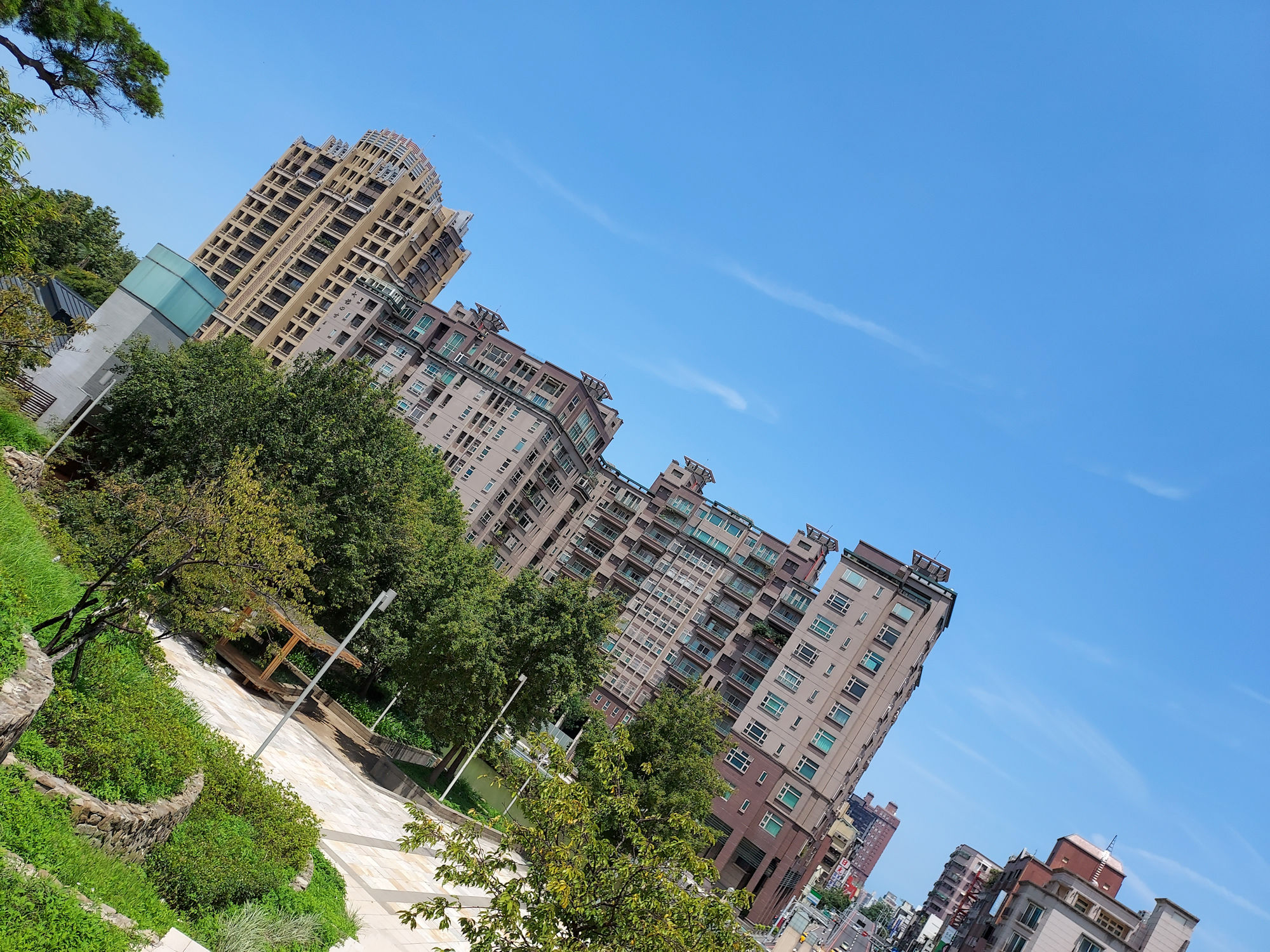}} \\
	\end{tabular}
	\caption{Real test images. The resolutions of these images are all 4000$\times$3000}. 
	\label{figure_real_images}
\end{figure*}
\begin{figure*}
	\centering
	\begin{tabular}[t]{cccc}
		\subfigure{\includegraphics[height=2.2cm]{Set1_04_small.jpg}} \hspace{0.1cm}
		\subfigure{\includegraphics[height=2.2cm]{Set2_01_small.jpg}} \hspace{0.1cm}
		\subfigure{\includegraphics[height=2.2cm]{Set5_01_small.jpg}} \hspace{0.1cm}
		\subfigure{\includegraphics[height=2.2cm]{Set8_01_small.jpg}} \\
		\subfigure{\includegraphics[height=3.57cm]{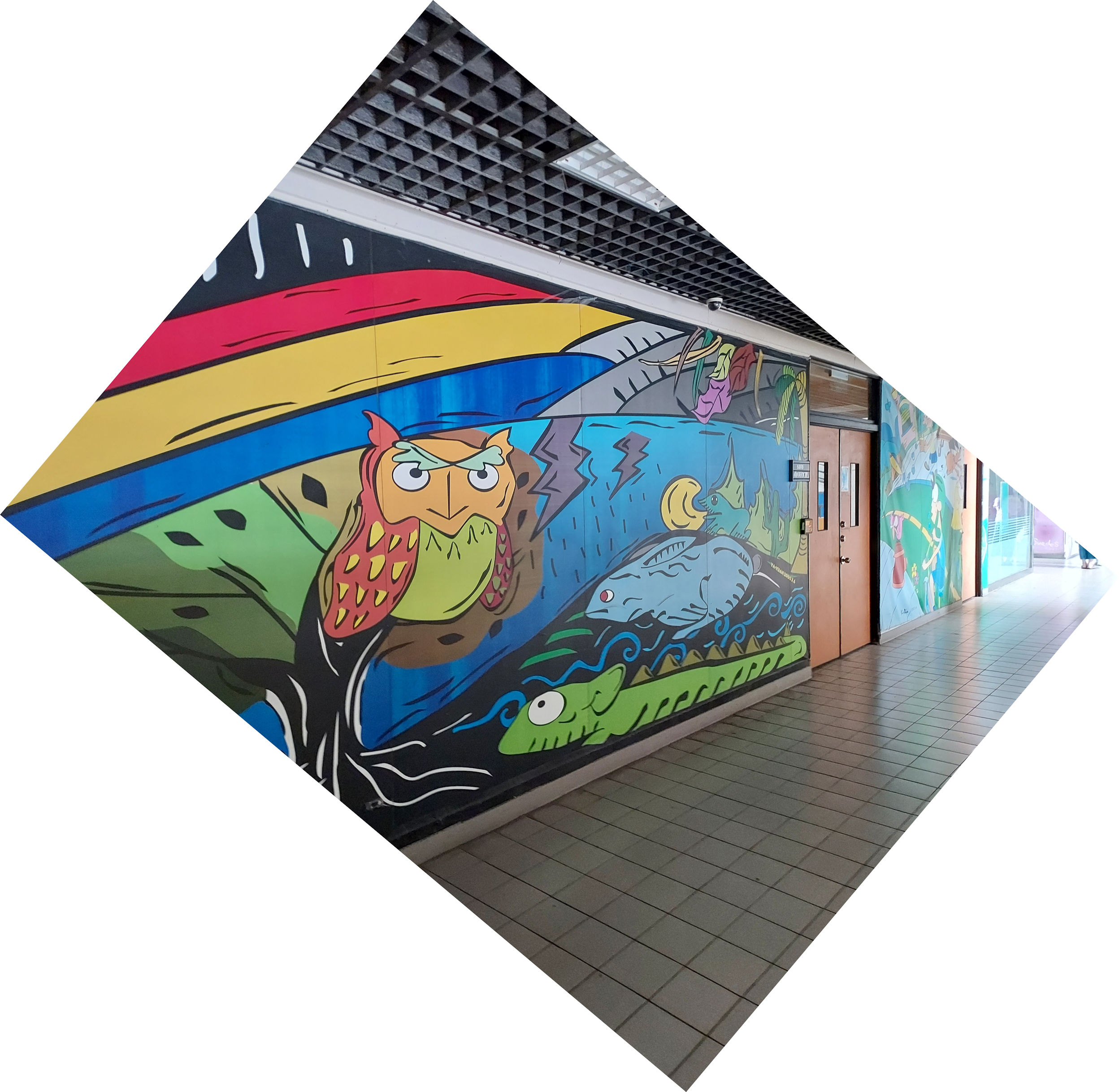}} \hspace{0.1cm}
		\subfigure{\includegraphics[height=3.64cm]{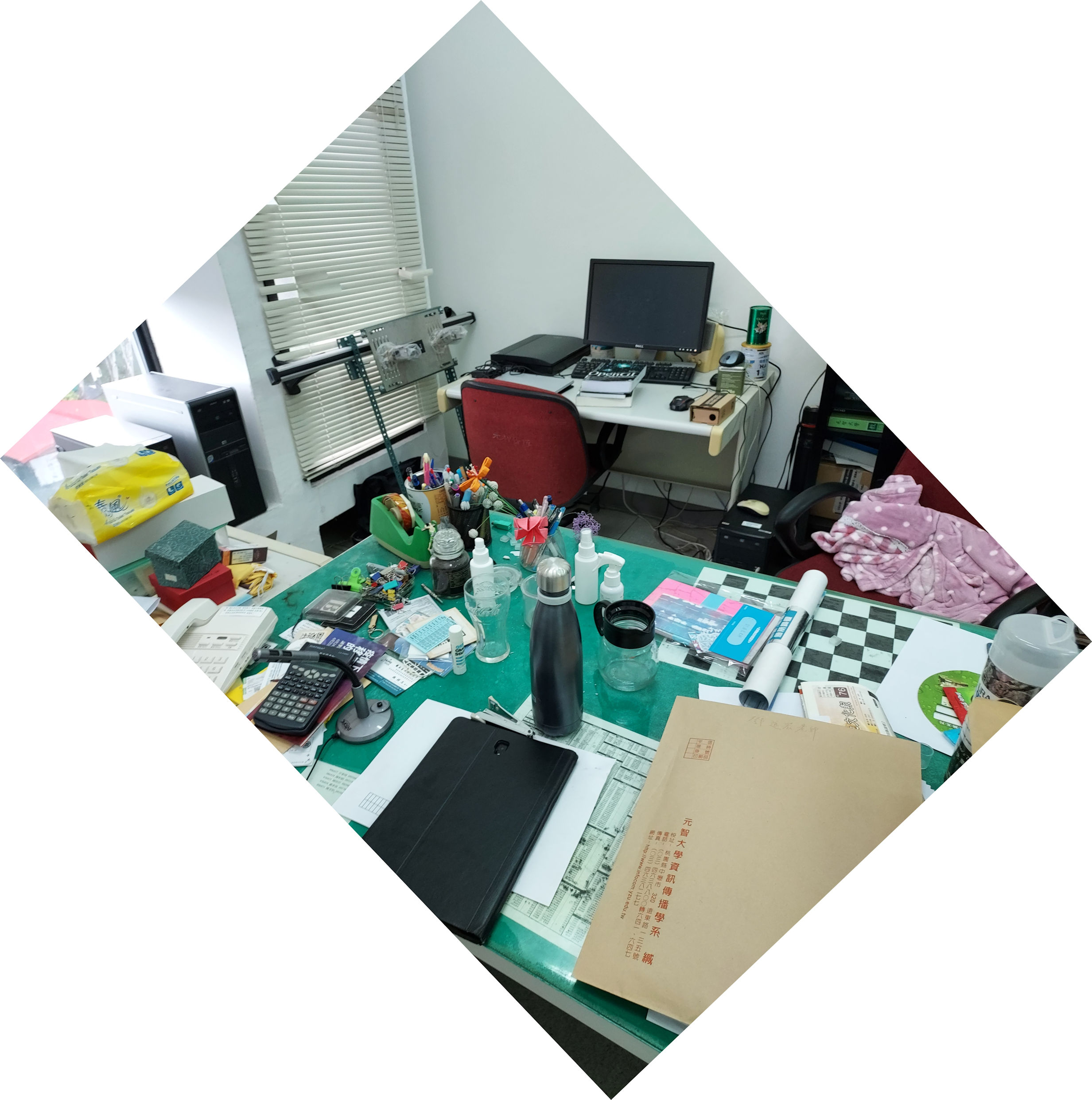}} \hspace{0.1cm}
		\subfigure{\includegraphics[height=3.52cm]{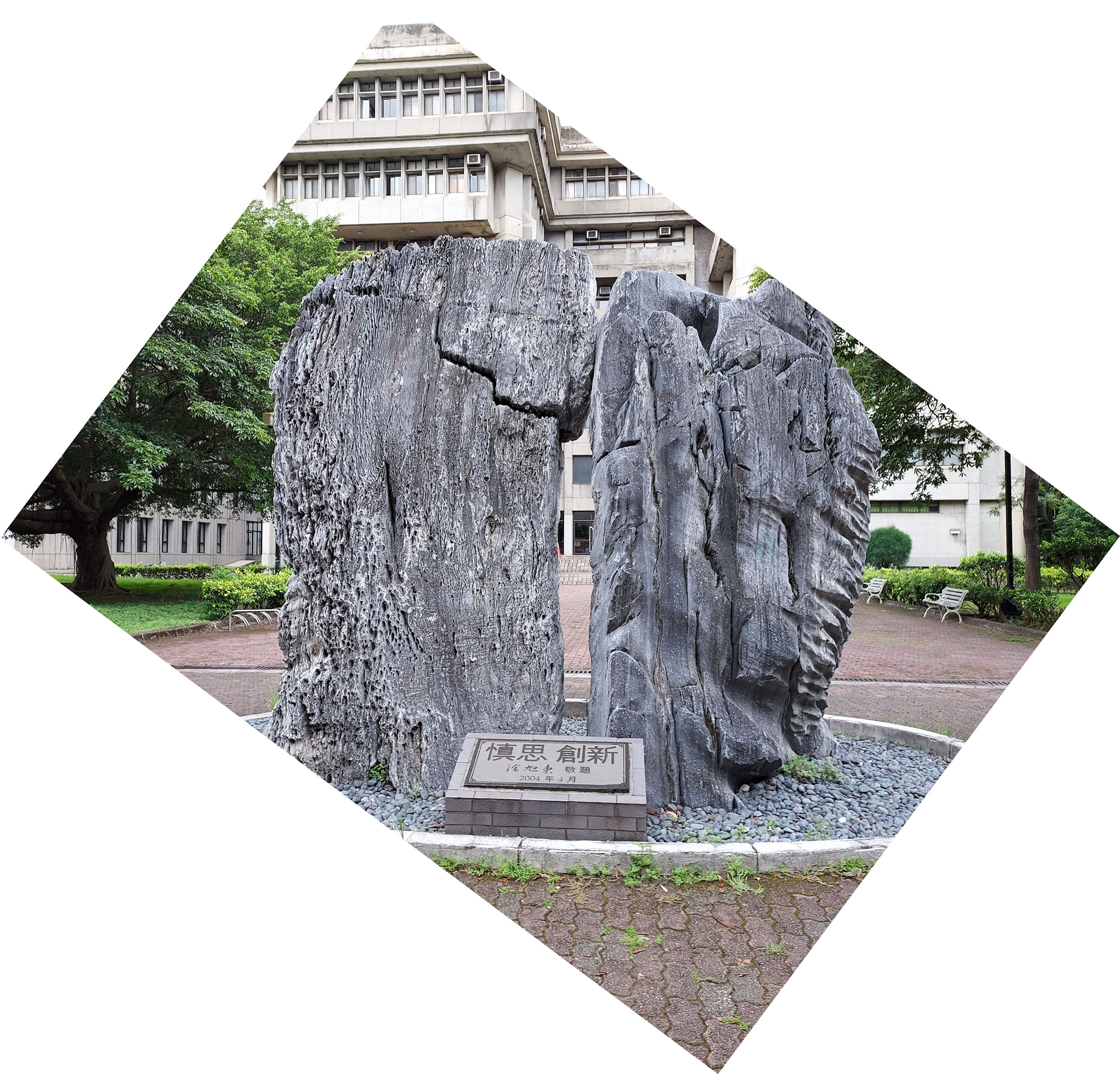}} \hspace{0.1cm}
		\subfigure{\includegraphics[height=3.49cm]{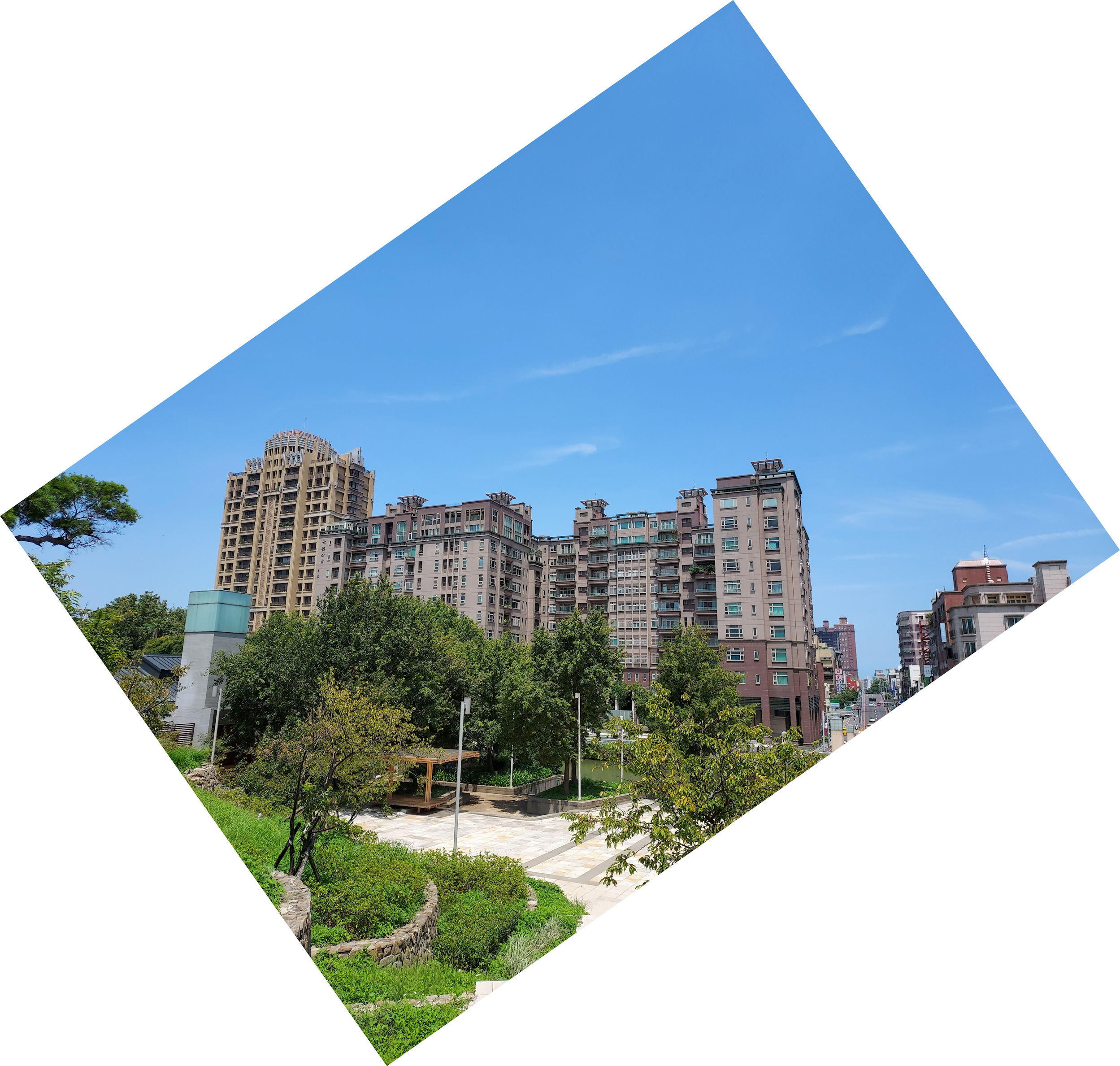}} \\
		\subfigure{\includegraphics[height=2.2cm]{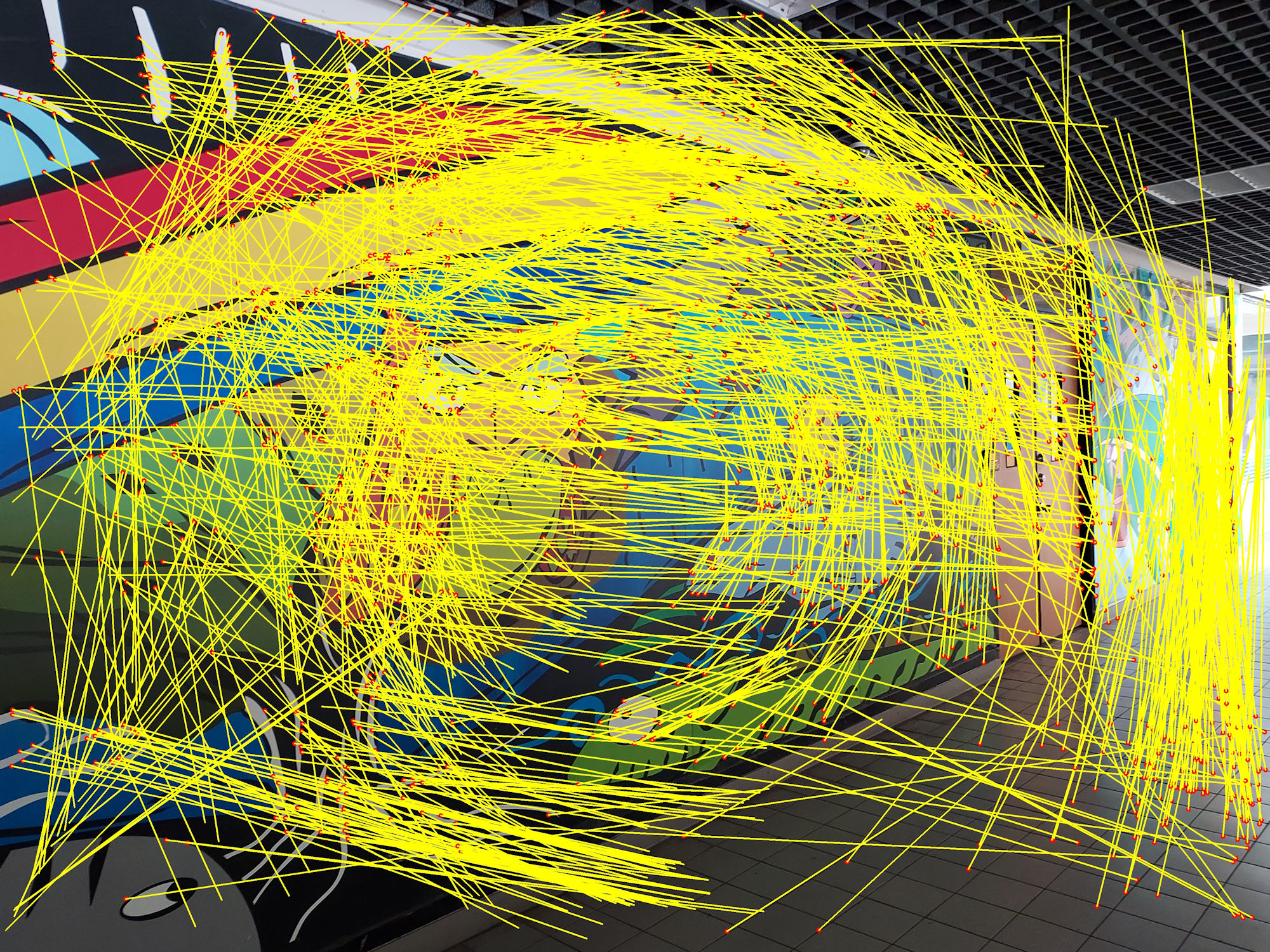}} \hspace{0.1cm}
		\subfigure{\includegraphics[height=2.2cm]{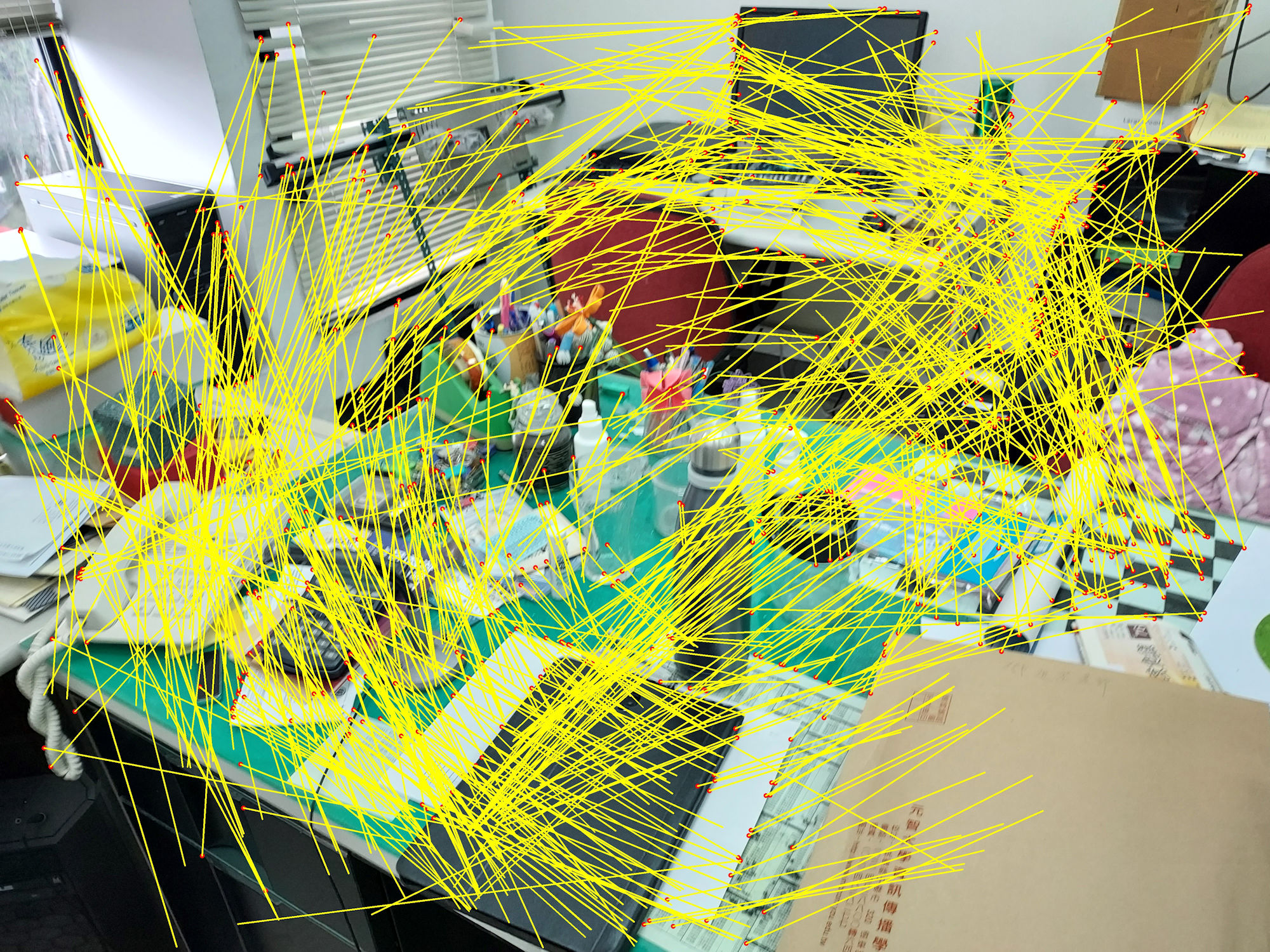}} \hspace{0.1cm}
		\subfigure{\includegraphics[height=2.2cm]{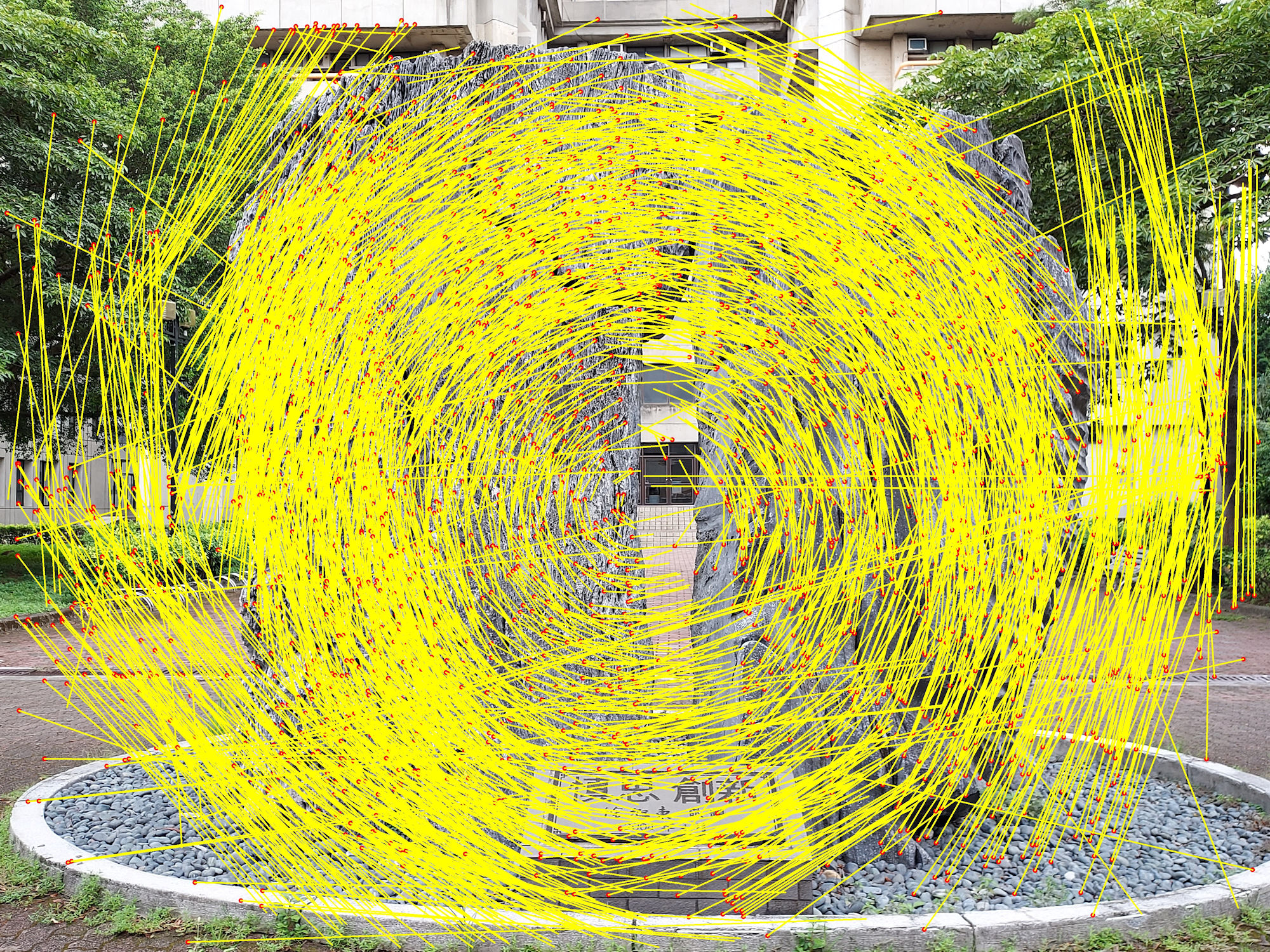}} \hspace{0.1cm}
		\subfigure{\includegraphics[height=2.2cm]{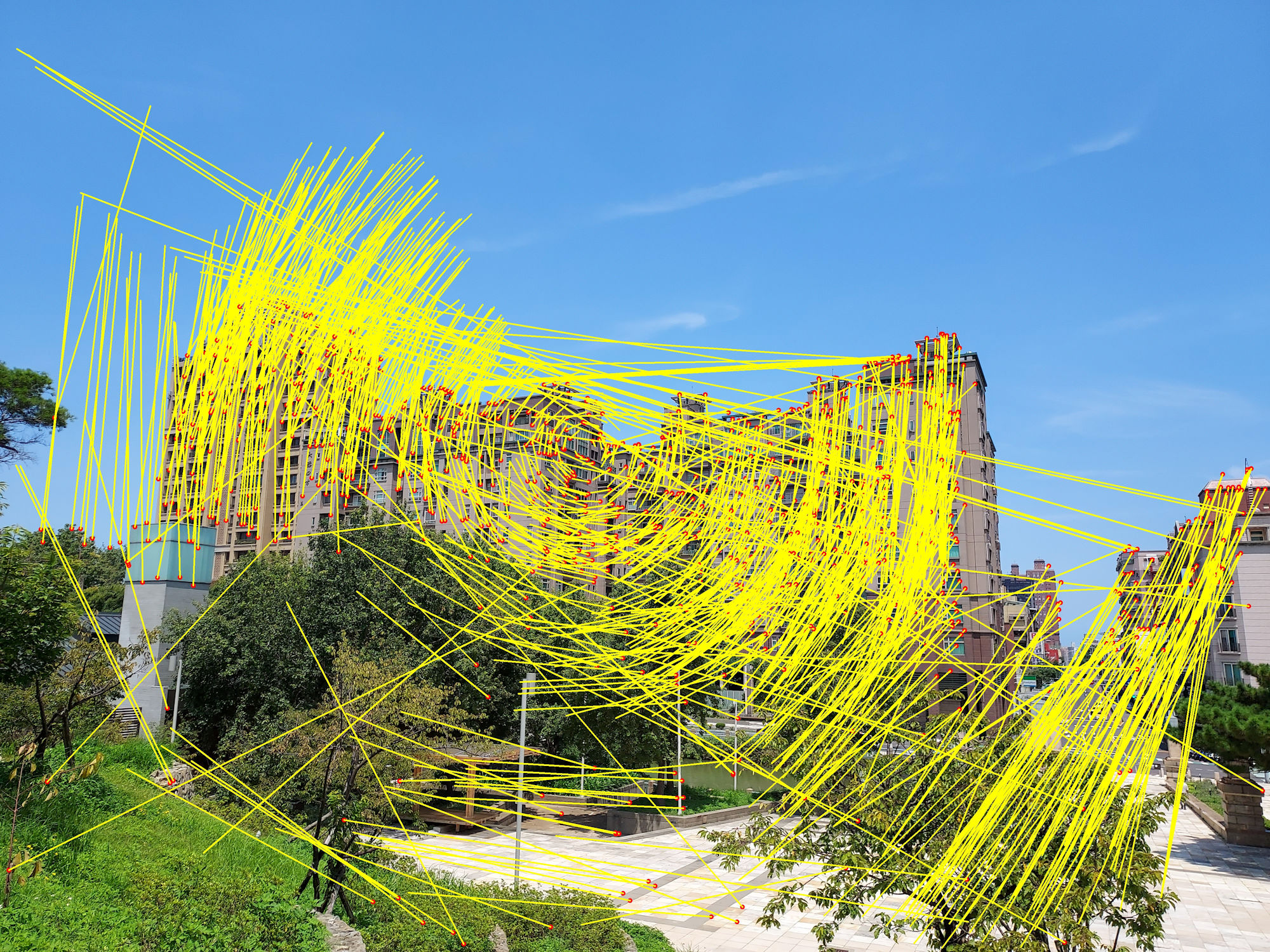}} \\
		\subfigure{\includegraphics[height=2.2cm]{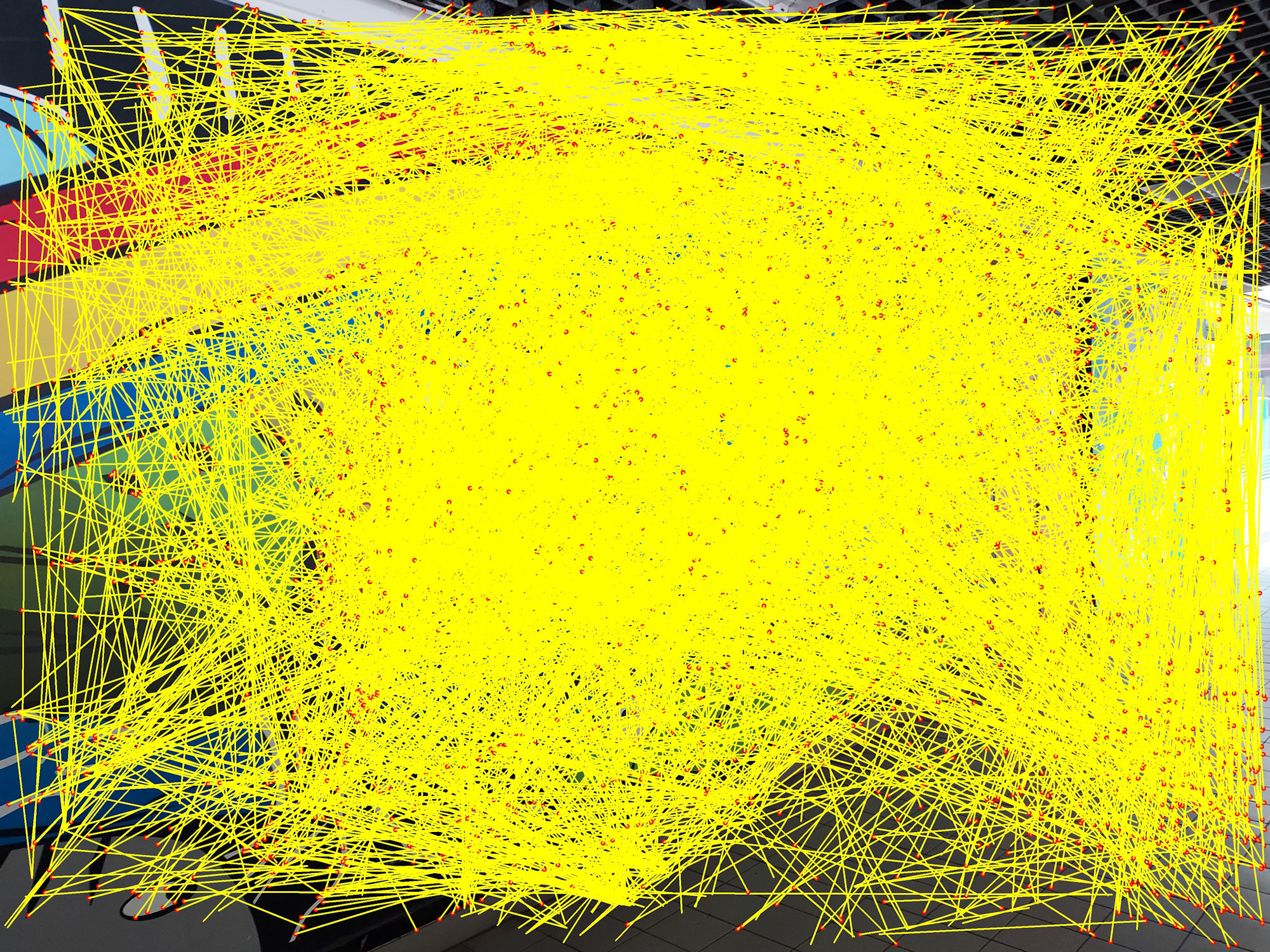}} \hspace{0.1cm}
		\subfigure{\includegraphics[height=2.2cm]{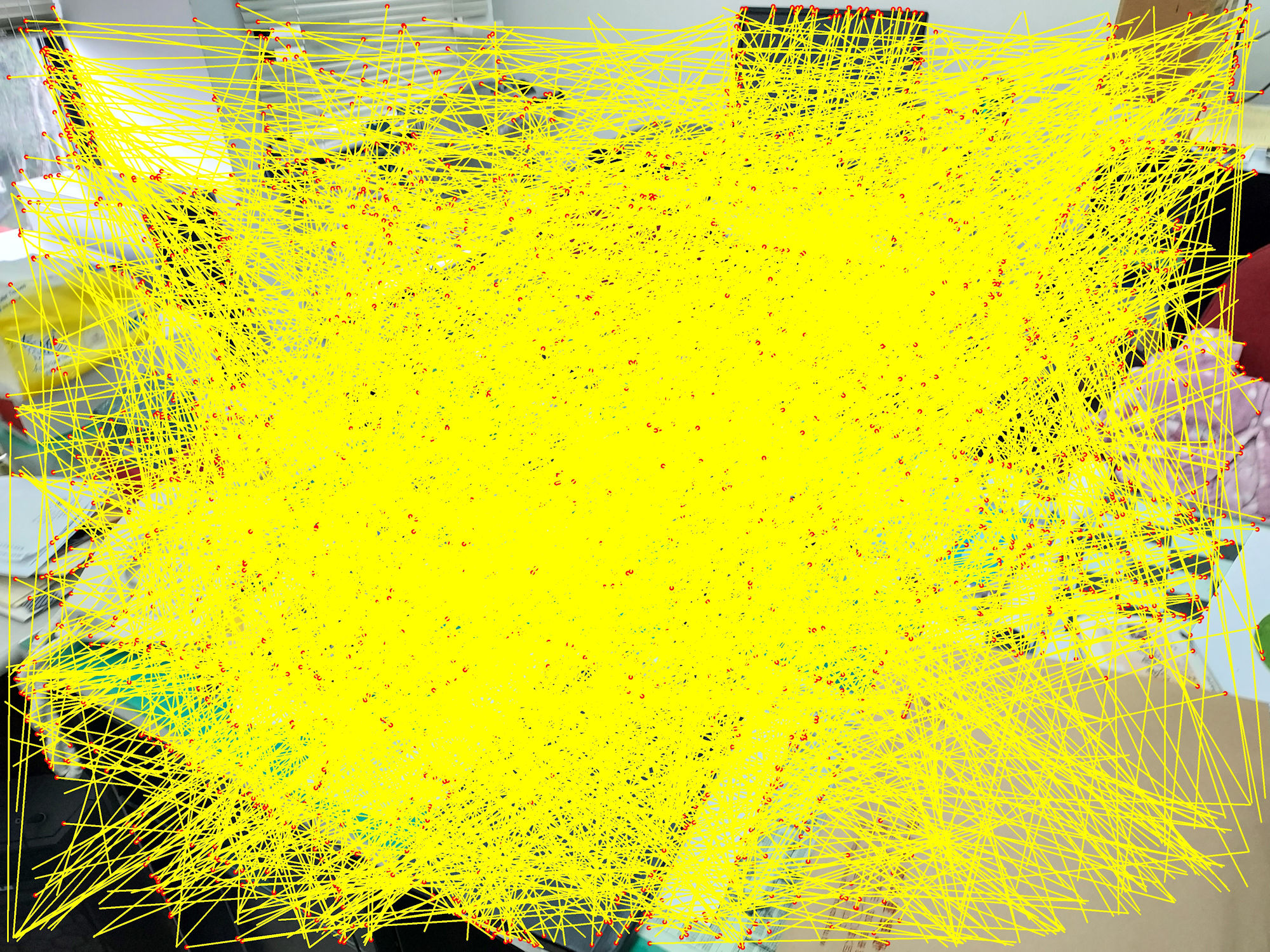}} \hspace{0.1cm}
		\subfigure{\includegraphics[height=2.2cm]{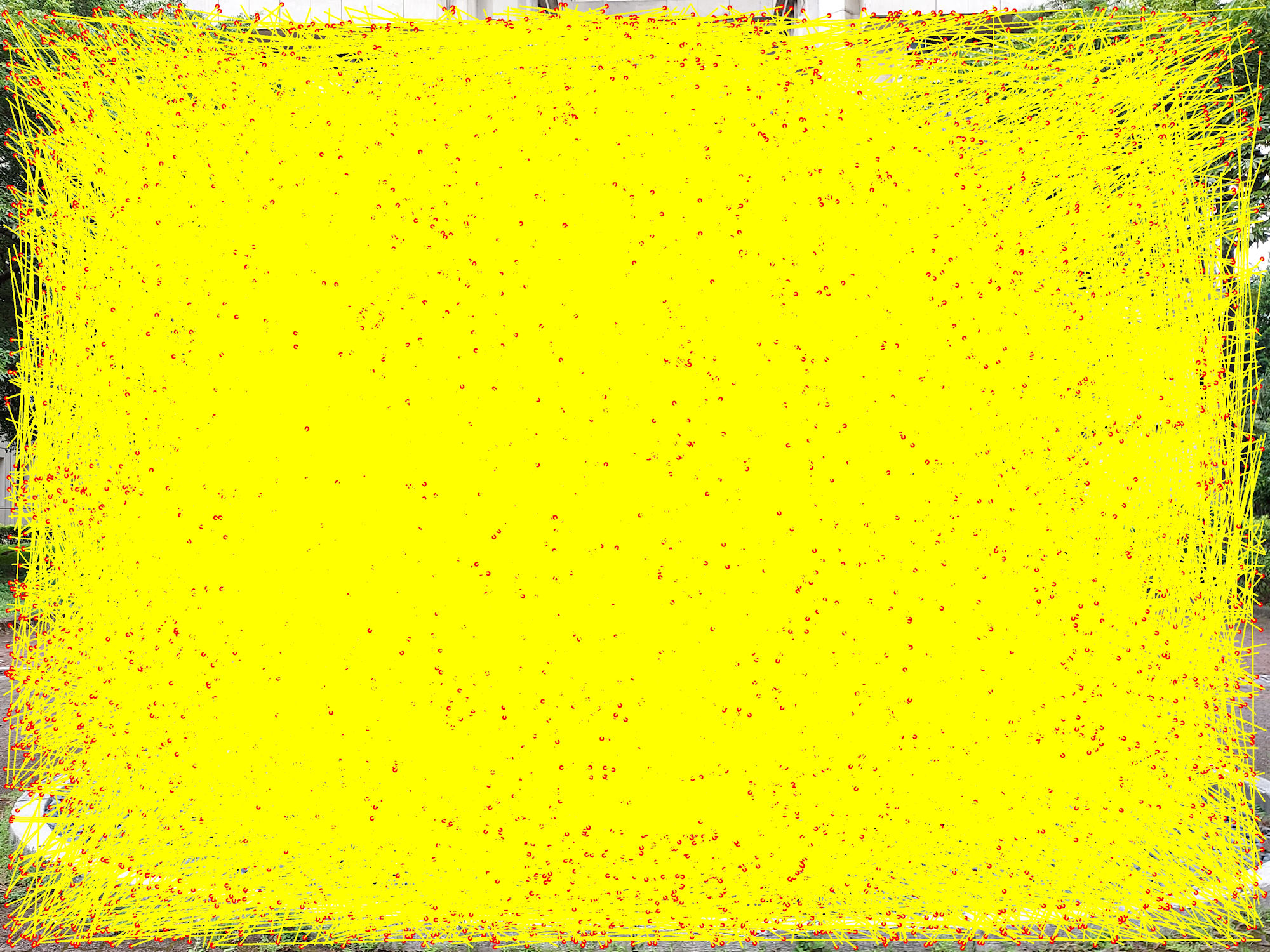}} \hspace{0.1cm}
		\subfigure{\includegraphics[height=2.2cm]{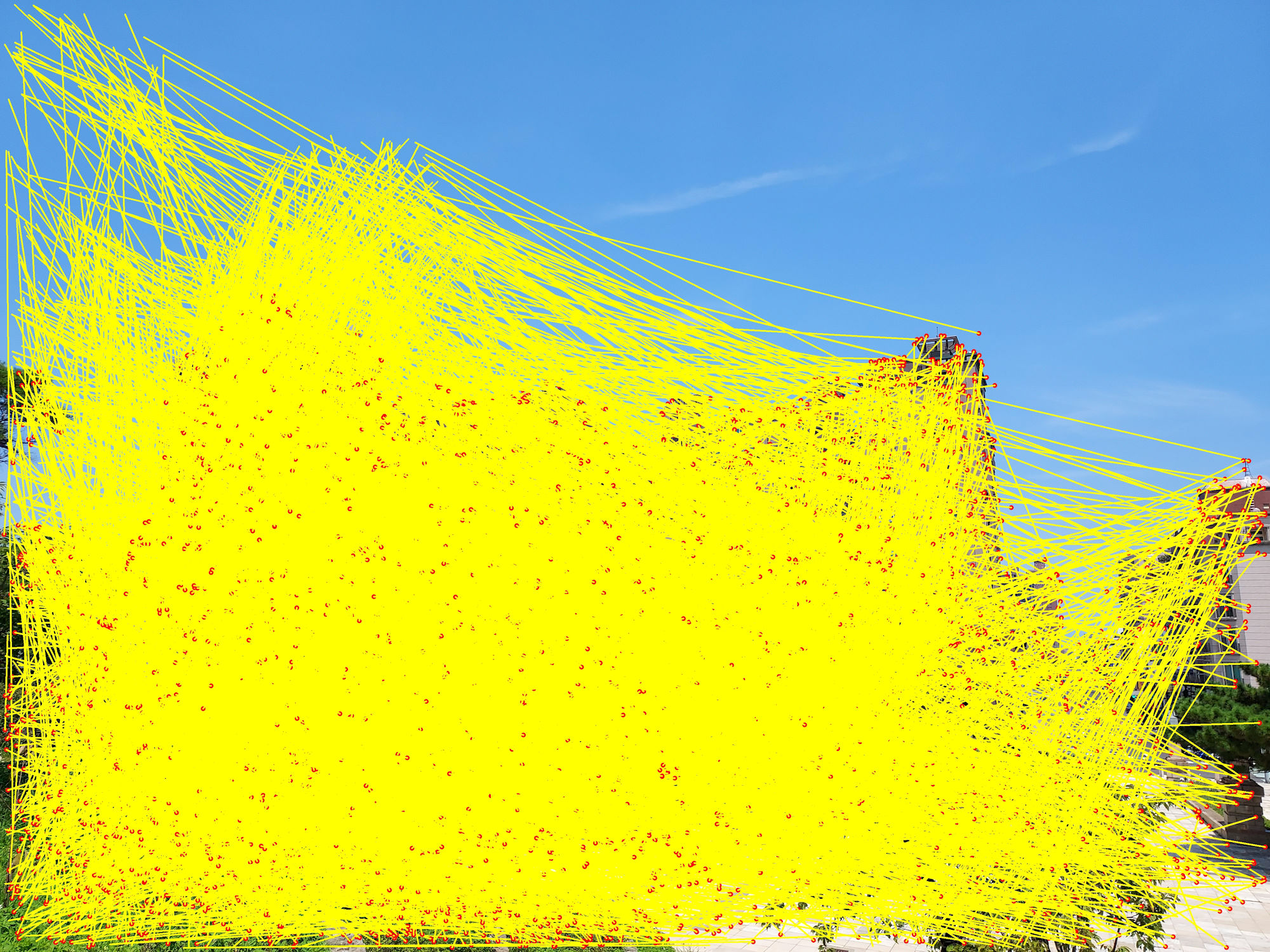}} \\			
	\end{tabular}
	\caption{Results of the real test images in Fig. \ref{figure_real_images}.}
	\label{figure_results_real_images}
\end{figure*}

\indent In addition to the aforementioned datasets, we also evaluated our framework on real images, as shown in Fig. \ref{figure_real_images}. Figure \ref{figure_results_real_images} shows the experimental results for these real images, where the first row shows the original first images and the second row shows the aligned second images. By examining these images, we observe that our image alignment can restore the second image to its proper orientation. Since for real images we do not know whether the match is correct or not, we can only visually display the feature displacement (\textit{i.e.}, the difference in coordinates of the corresponding points) in the figure. The third row of Fig. \ref{figure_results_real_images} shows the results of our method (SO+EPI+IA), where each yellow line represents the movement of a feature point. Since these images contain rotations around the image view direction, the resulting motion field should present a swirling shape. A swirl-like motion field can indeed be observed in the images in the third row of Fig. \ref{figure_results_real_images}. Furthermore, there exist some incorrect matches where the yellow lines possess random orientation. One reason for these incorrect matches is that the beginning stage our framework does not have the spatial order and epipolar geometry models and so feature matching relies merely on the similarity of feature descriptors, which easily generate incorrect matches. In fact, if we employ the subsequent spatial order and epipolar geometry models to examine the matched features created at the beginning stage, there is a chance that these incorrect matches can be removed. The fourth row of Fig. \ref{figure_results_real_images} shows the matching results obtained by the brute force approach. From these images, we can see that the results are very messy and a swirling shape cannot be observed, indicating that there are many incorrect matches. \\
\indent To further examine the test results of real images, we list some performance indices of our method and the brute force approach in Table \ref{table_result_data_real_images}. Since we do not have the ground truth of these real images, we use RANSAC with epipolar geometry to determine the inliers/outliers and use the data to determine the matching precision. Although it is not possible to fully determine the correctness of a set of correspondences, the data can still provide a reference for the matching precision. As can be seen from the table, in comparison with the brute force approach, our method achieves much higher matching precisions and requires much less matching time. \\
\begin{table*}
	\footnotesize
	\centering
	\caption{Performances for the images in Fig. \ref{figure_real_images}.}
	\begin{tabular}{p{3cm}ccccc} 
		\hline
		Test images & Approach &  Fig. \ref{figure_real_images}(a) & Fig. \ref{figure_real_images}(b) & Fig. \ref{figure_real_images}(c) & Fig. \ref{figure_real_images}(d) \\ \hline
		\# Features in view 1 &                     & 47,844  & 27,865  & 136,740 & 88,498  \\ 
		\# Features in view 2 &                     & 48,225  & 33,789  & 132,759 & 81,397 \\ \hline
		\multirow{2}{3cm}{\# Correspondences}  & BF & 33,946  & 13,966  & 52,646  & 29,904 \\ 
		& SO+EPI+IA   & 13,195  & 3,378   & 12,451  & 5,020 \\ \hline
		\multirow{2}{3cm}{Precision$^a$}       & BF & 1.71\%  & 4.04\%  & 12.98\% & 11.20\% \\ 
		& SO+EPI+IA   & 73.76\% & 68.35\% & 89.78\% & 84.80\% \\ \hline
		\multirow{2}{1cm}{Matching time (s)}   & BF & 57.54   & 26.28   & 681.63  & 263.41 \\  
		& SO+EPI+IA   & 3.03    & 1.44    & 23.13   & 9.95 \\ \hline
	\end{tabular}\\
	\footnotesize{$^a$ The precision is estimated based on epipolar geometry.}
	\label{table_result_data_real_images}
\end{table*}
\indent The experimental results for the real images indicate that our framework is feasible for practical applications. In particular, our framework is designed to provide a more efficient way of matching the features of two images. Since the matching is still based on the similarity of descriptors, it cannot guarantee that all matches are correct. Our framework is mainly used to establish the initial feature matching between two images quickly. It still requires some other robust methods such as RANSAC + epipolar geometry or vector field consensus \cite{Ma14} to remove incorrect matches. Nevertheless, our method is effective in increasing the proportion of inliers during the initial feature matching phase. This increased inlier rate will be beneficial for subsequent outlier removal approaches, such as RANSAC which requires sampling on the data. The improved inlier rate will significantly increase the computational efficiency of RANSAC as well as the chance of finding an accurate model.

\section{Conclusion}
\label{section_conclusion}
\indent In this paper, we proposed a progressive image feature matching framework based on the spatial order of feature points. This framework allows feature matching to be limited to a specific area, thus significantly improving computational efficiency. Epipolar geometry can be integrated into this framework to restrict the matching to a much smaller area, thus further enhancing matching efficiency. Furthermore, by decomposing the fundamental matrix of epipolar geometry, we proposed an image alignment method that can solve the problem of image rotation induced by the spatial order model, making our system more suitable for practical applications. In fact, our framework can be seen as a machine learning system. By learning a model from previously matched features, the learned model can be effectively used to regulate subsequent matches. We conducted a series of experiments using a standard benchmark dataset, self-generated simulated images, and real images. The results demonstrated that our framework has the potential to be an outstanding strategy for image feature matching. \\
\indent A notable point is that our framework does not limit the feature detection and descriptor generation techniques used. Therefore, by using a more robust feature detection and descriptor technique, we can theoretically make our method more robust for images with large deformation. In addition, since our system is a guided feature matching technique on image space, it can be combined with other descriptor-based speed-up methods, such as approximate nearest neighbors or cascade hashing, to achieve even more efficient feature matching. Moreover, since our framework treats each feature point individually, it can be implemented in parallel. \\
\indent In practice, if the epipolar geometry does not hold, our framework can still use the spatial order model alone to filter the feature points. However, in such cases, the problem of image rotation should be dealt with using other techniques. In future, we will try to combine the proposed framework with other descriptor-based searching strategies to further enhance matching efficiency while preserving matching accuracy.

\section*{Acknowledgements}
This work was supported by the National Science and Technology Council, Taiwan, under Grant No. MOST 106-2221-E-155-067.

	\makeatother
	
	\setBibStyle
	\bibliography{bibliography/Reference.bib}
\end{document}